\documentclass{article}
\usepackage{amsmath}
\usepackage{amsfonts}
\usepackage{amssymb}
\usepackage{graphicx}
\usepackage[english]{babel}
\usepackage{hyperref} 
\usepackage{enumitem} 
\usepackage{natbib}
\usepackage{tabularx}
\usepackage{subcaption} 
\usepackage{listings}
\usepackage{xcolor}
\usepackage{threeparttable} 
\usepackage{subcaption}
\usepackage{threeparttable}

\definecolor{codegreen}{rgb}{0,0.6,0}
\definecolor{codegray}{rgb}{0.5,0.5,0.5}
\definecolor{codepurple}{rgb}{0.58,0,0.82}
\definecolor{backcolour}{rgb}{0.95,0.95,0.92}
\lstdefinestyle{mystyle}{
    backgroundcolor=\color{backcolour},   
    commentstyle=\color{codegreen},
    keywordstyle=\color{magenta},
    numberstyle=\tiny\color{codegray},
    stringstyle=\color{codepurple},
    basicstyle=\ttfamily\footnotesize,
    breakatwhitespace=false,         
    breaklines=true,                 
    captionpos=b,                    
    keepspaces=true,                 
    numbers=left,                    
    numbersep=5pt,                  
    showspaces=false,                
    showstringspaces=false,
    showtabs=false,                  
    tabsize=2
}
\usepackage{booktabs}
\usepackage{longtable}
\usepackage{multirow}
\graphicspath{ {./figures/} }
\usepackage[margin=1in]{geometry}
\usepackage{babel,blindtext}

\newcommand{\citeg}[1]{\text{}}

\title{LingBench++: A Linguistically-Informed Benchmark and Reasoning Framework for Multi-Step and Cross-Cultural Inference with LLMs}

\author{
    Da-Chen Lian$^{1}$, 
    Ri-Sheng Huang$^{2}$, 
    Pin-Er Chen$^{1}$, 
    Chunki Lim$^{1}$, 
    You-Kuan Lin$^{3}$, \\
    Guan-Yu Tseng$^{1}$, 
    Tzu-Cheng Yang$^{2}$, 
    Zhen-Yu Lin$^{4}$,
    Pin-Cheng Chen$^{4}$, 
    Shu-Kai Hsieh$^{1}$ \\
    \\
    $^{1}$Graduate Institute of Linguistics, National Taiwan University \\
    $^{2}$Department of Computer Science and Information Engineering, National Taiwan University \\
    $^{3}$Department of Electrical Engineering, National Taiwan University \\
    $^{4}$Department of Foreign Languages and Literatures, National Taiwan University
}
\date{\today}

\begin{document}

\maketitle

\begin{abstract}

We propose LingBench++, a linguistically-informed benchmark and reasoning framework designed to evaluate large language models (LLMs) on complex linguistic tasks inspired by the International Linguistics Olympiad (IOL). Unlike prior benchmarks that focus solely on final answer accuracy, LingBench++ provides structured reasoning traces, stepwise evaluation protocols, and rich typological metadata across over 90 low-resource and cross-cultural languages. We further develop a multi-agent architecture integrating grammatical knowledge retrieval, tool-augmented reasoning, and deliberate hypothesis testing. Through systematic comparisons of baseline and our proposed agentic models, we demonstrate that models equipped with external knowledge sources and iterative reasoning outperform single-pass approaches in both accuracy and interpretability. LingBench++ offers a comprehensive foundation for advancing linguistically grounded, culturally informed, and cognitively plausible reasoning in LLMs.

\end{abstract}

\section{Introduction}

The International Linguistics Olympiad (IOL) presents uniquely challenging problems that require solvers to induce linguistic rules from micro-data, often in low-resource or unfamiliar languages. These problems test not just surface-level pattern recognition but demand multi-step abstraction, structural reasoning, and cultural inference.\footnote{\url{https://ioling.org/}} Unlike many other olympiads, the Linguistics Olympiad does not require any prior knowledge of the languages involved. All problems are designed to be self-contained, allowing participants to discover underlying linguistic rules purely through logical reasoning and pattern analysis~\citep{bozhanov-derzhanski-2013-rosetta}.

While large language models (LLMs) such as GPT-4 and Gemini have achieved strong performance on many reasoning-related tasks, their ability to solve IOL-style problems---especially those involving multimodal symbols, rare scripts, or typological diversity---remains underexplored.


Linguistics problems in IOL are essentially logic puzzles, usually based on real languages, allowing for the discovery of specific linguistic phenomena through logical reasoning. Problems typically consist of a corpus (dataset) in an unfamiliar language accompanied by its English translations (either ordered or unordered). Solving the problem requires logical thinking and attention to detail to decode certain aspects of the language, such as the meaning of certain words or some grammatical rules. All linguistic problems are internally consistent (self-consistent) and require no additional knowledge (self-sufficient). This means that once certain rules are discovered, they will apply to all examples in the problem (though some rules may have exceptions), and all information needed to solve the problem is found within the problem itself.

\begin{figure}
\includegraphics[scale=0.3]{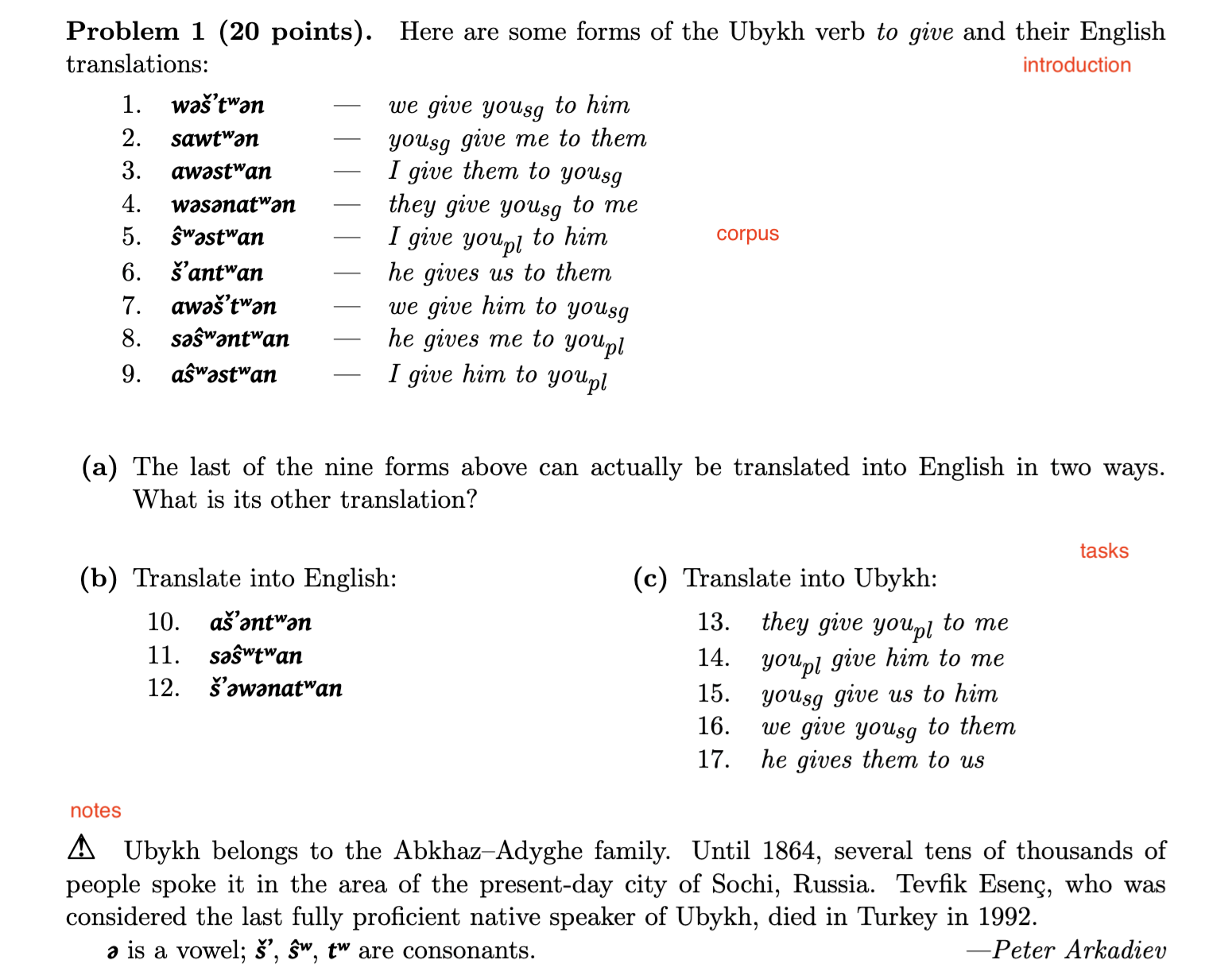}
\caption{Example of an IOL Problem}\label{iol}
\end{figure}

Every linguistics problem is structured into four parts: an introduction, a corpus, tasks, and notes. An example is shown in Figure~\ref{iol}.

\begin{itemize}
    \item \textbf{Introduction:} Provides information about the language featured in the problem. If the introduction is complex and contains additional information, this information is likely to be relevant to solving the problem.
    \item \textbf{Corpus:} Contains the examples based on which the tasks should be solved.
    \item \textbf{Tasks:} Follow the corpus and typically include ``Determine the correct correspondences,'' ``Translate into English,'' and ``Translate into [...],'' among others.
    \item \textbf{Notes:} Provide data about the language featured in the problem, relevant phonetic information, and details about specific words.
\end{itemize}

In addition to abstract linguistic reasoning, many IOL problems incorporate elements that go beyond standard textual input, requiring models to process non-standard scripts, phonetic transcriptions, and visual symbol systems. Some problems involve rare or extinct writing systems---occasionally ones not yet fully encoded in Unicode---demanding the recognition and manipulation of unfamiliar glyphs~\citep{shih2025reasoningglyphsevaluationllms}. Others rely on International Phonetic Alphabet (IPA) representations, tone contour symbols, or constructed orthographies that encode morphophonemic information. A subset of tasks also includes pictographic cues, spatial arrangements, or logical diagrams, which are essential to deciphering underlying rules.

\begin{figure}[!htb]
    \centering
    \begin{subfigure}{0.42\textwidth}
        \includegraphics[width=\linewidth]{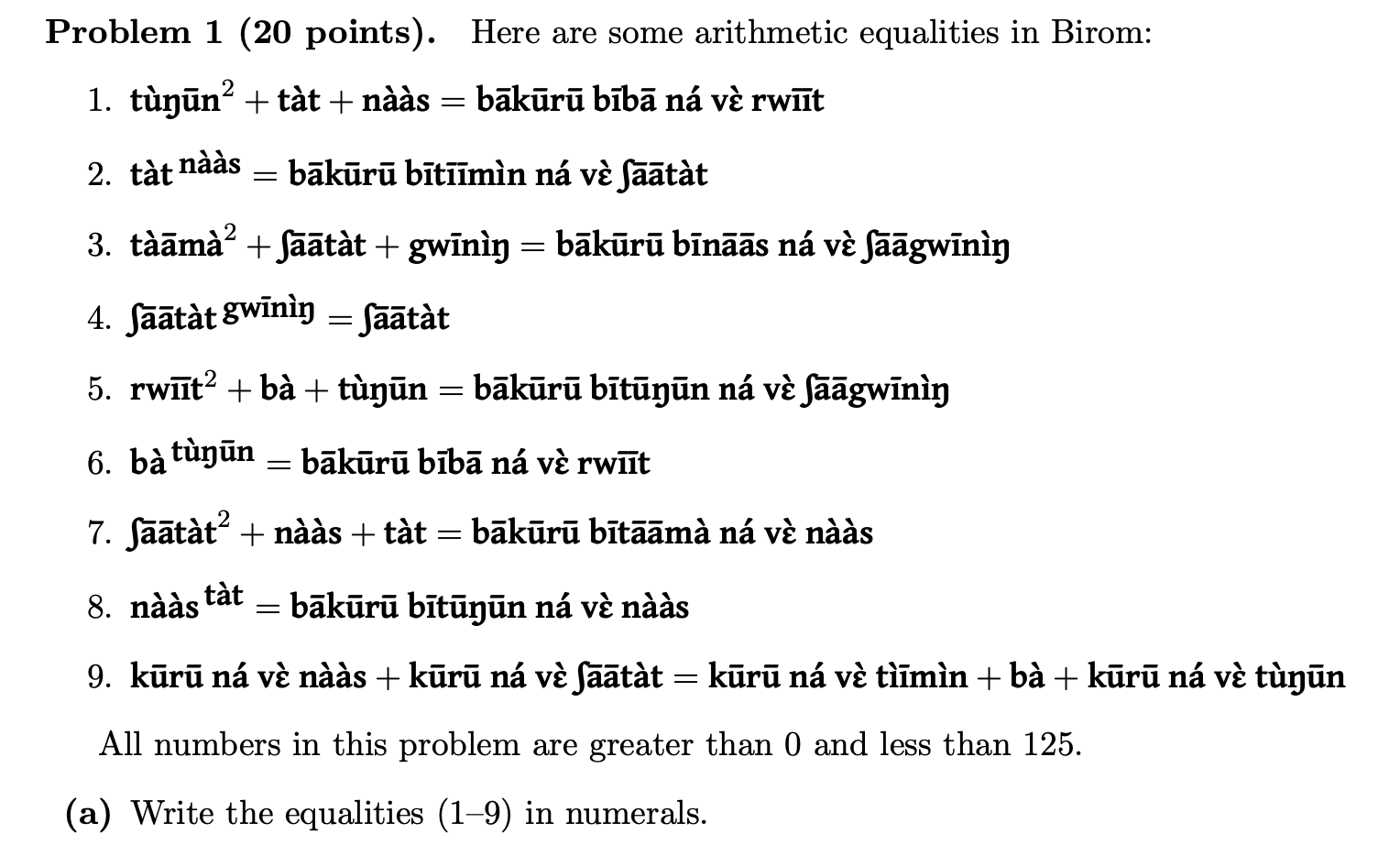}
        \label{fig:2017-1}
    \end{subfigure}\hfill
    \begin{subfigure}{0.32\textwidth}
        \includegraphics[width=\linewidth]{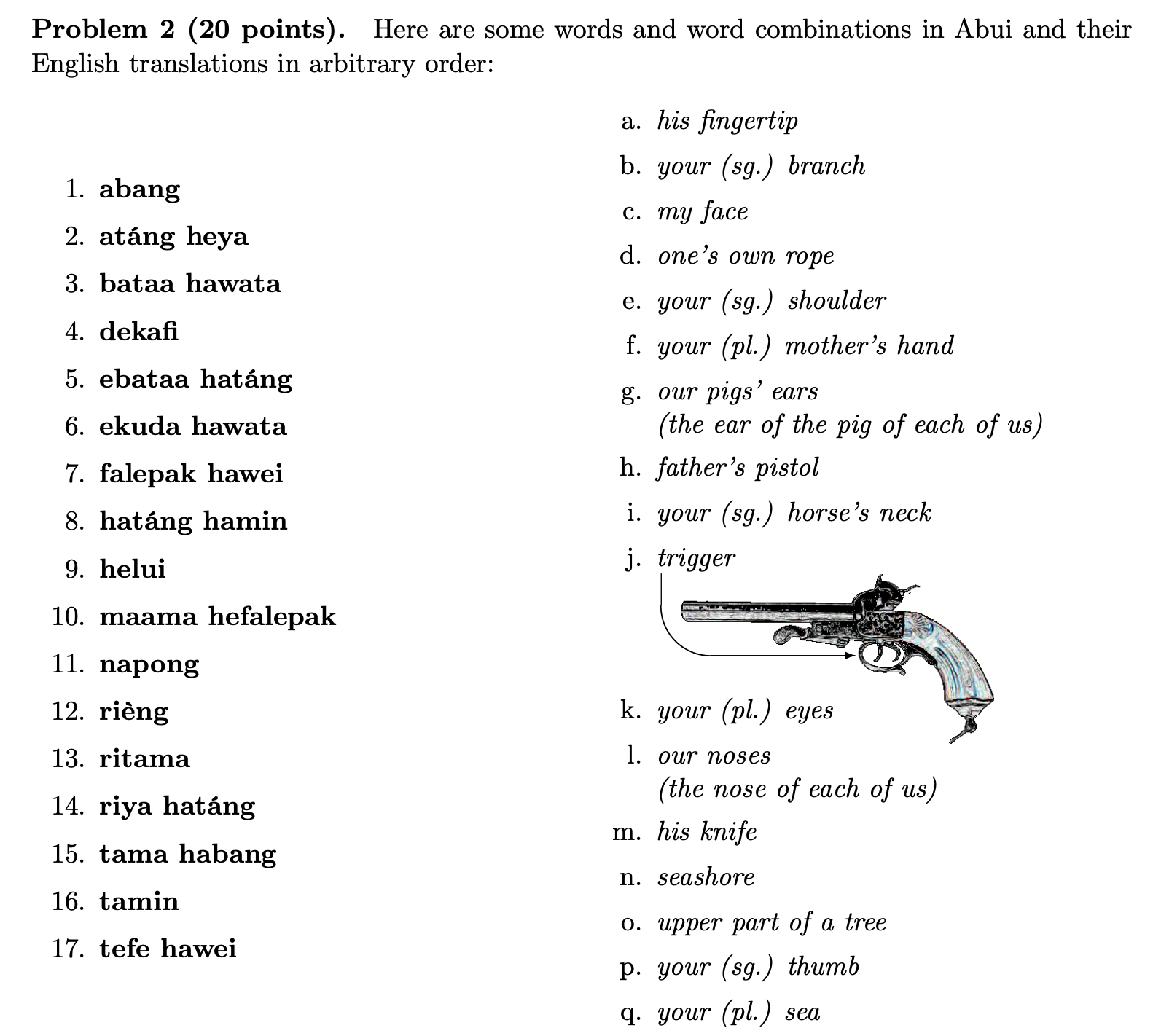}
        \label{fig:2017-2}
    \end{subfigure}\hfill
    \begin{subfigure}{0.32\textwidth}
        \includegraphics[width=\linewidth]{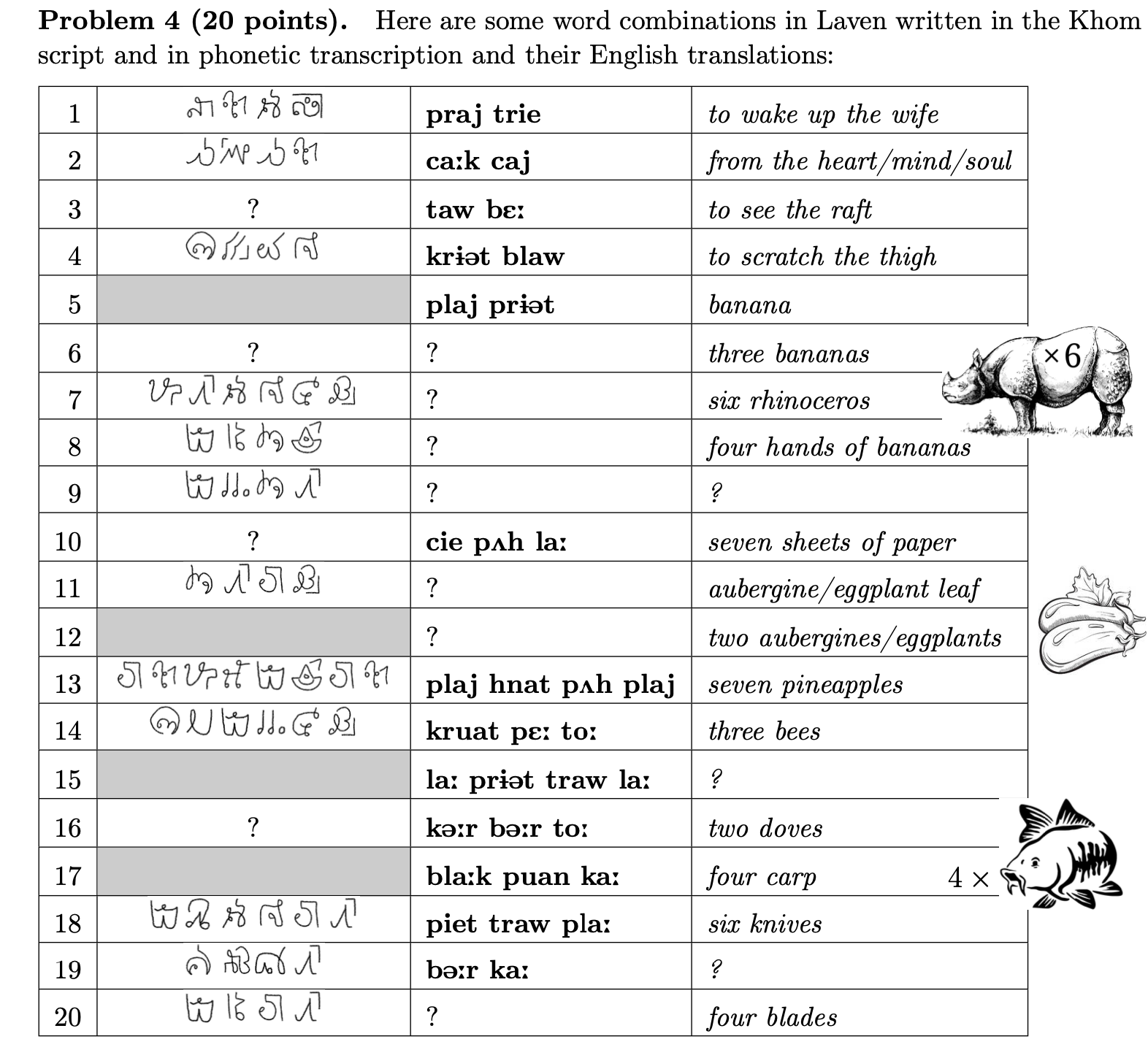}
        \label{fig:2017-4}
    \end{subfigure}
    \caption{Three Problems in IOL 2017}
    \label{fig:2017_problems}
\end{figure}

These multimodal components challenge LLMs not only in terms of language understanding, but also in visual parsing, symbol grounding, and cross-modal inference (see Figure~\ref{fig:2017_problems}). While recent vision-language models (e.g., GPT-4V, Gemini, Kosmos-2) have made progress in processing images and text jointly, their ability to integrate these modalities in service of structured linguistic reasoning remains limited. IOL tasks thus offer a compelling testbed for assessing and advancing multimodal reasoning models, where the interplay between text, symbol, and visual structure is critical to problem solving.

One distinctive aspect of IOL problems lies in their \textbf{cross-cultural and semantic depth}. Beyond the structural reasoning over phonology, morphology, and syntax, many tasks require solvers to engage with \emph{low-resource or endangered languages} under conditions of \emph{micro-data}. More significantly, some problems explicitly involve semantic inference, cultural conceptualization, or sociolinguistic reasoning---for instance, deciphering kinship terms, numeral systems, metaphorical extensions, or culturally situated deixes. These tasks compel both human and AI solvers to imagine how meaning might be constructed in unfamiliar cultural worlds, often requiring \emph{cross-linguistic abstraction} or \emph{anthropological imagination}. For LLMs, this poses a profound challenge: it tests their ability to generalize across not only linguistic structures but also cognitive and cultural domains. IOL problems, therefore, serve not only as puzzles of language form but as tests of situated meaning-making and cultural flexibility, offering a rigorous probe into the limits of LLMs' representational and interpretive capacity across diverse human experiences.

These complex challenges expose the limitations of current LLMs and existing evaluation methods, which often prioritize final-answer accuracy over the reasoning process. To address these gaps, this paper introduces LingBench++ and makes three primary contributions. First, we detail the construction of the LingBench++ benchmark, featuring 96 IOL problems with expert-verified reasoning traces and rich typological metadata, and we advocate for a shift towards stepwise evaluation protocols to better diagnose model reasoning (Section~\ref{sec:LOBSTER}). Second, to contextualize the challenges for LLMs in handling linguistic diversity, we conduct an empirical study of a state-of-the-art model's multilingual capabilities, analyzing its performance with respect to resource availability and typological features (Section~\ref{section:utilizing_ref_grammar}). Finally, as a step toward more robust linguistic reasoning, we propose and present preliminary experiments on a multi-agent framework that leverages external grammatical knowledge and iterative hypothesis testing to solve IOL problems (Section~\ref{section:multiagentframework}).

\section{Overview of Linguistic Reasoning}
\subsection{Large Reasoning Models}


Large reasoning models represent a class of AI systems designed to perform complex, structured problem-solving beyond basic pattern recognition. While recent advances in large language models (LLMs) have revolutionized natural language understanding and generation, significant challenges persist in achieving robust reasoning capabilities—particularly for tasks requiring multi-step abstraction, symbolic verification, and constraint-based hypothesis testing. These limitations are especially evident in domains demanding structural inference from minimal data, such as linguistic analysis where practitioners deduce grammatical patterns from sparse exemplars in unfamiliar languages.

To address these gaps, several reasoning-enhancement paradigms have emerged:
\begin{itemize}
    \item Chain-of-Thought (CoT) Prompting: Improves basic stepwise reasoning by generating intermediate rationales but lacks systematic verification mechanisms, limiting effectiveness on structurally complex problems \citep{kojima2023zeroshot, wei2023chainofthought, zhou2023leasttomost}.

    \item Tree-of-Thoughts (ToT): Extends CoT by enabling parallel hypothesis exploration, backtracking, and state evaluation, enhancing capabilities for combinatorial rule induction \citep{long2023tot, ranaldi-2024-tree, yao2023tot}.
    
    \item Hybrid Tool-Integrated Approaches: Frameworks like ReAct (reasoning + action), Reflexion (self-correcting reasoning), and Toolformer combine LLMs with external tools (calculators, code interpreters) and refinement cycles for error correction and tool-augmented inference \citep{he2025selfcorrection, gao2025embedding, paranjape2023multistepreasoning, schick2023toolformer, wu2025chainoftool}.

    \item Multimodal Architectures: Models such as GPT-4V and Gemini incorporate visual, tabular, and symbolic inputs but show inconsistent performance when processing aligned modalities; such as phonetic charts or morphosyntactic paradigms in linguistics \citep{qi2023geminivsgpt4v, he2025selfcorrection, zhang2024multimodalcot}.
\end{itemize}
These models shift LLMs from passive text generators toward deliberative, agentic systems capable of exploring solution spaces. Their efficacy is frequently evaluated through human-level reasoning benchmarks like the International Linguistics Olympiad (IOL) \citep{sahin-etal-2020-puzzling, chi-etal-2024-modeling}, where success requires inferring linguistic structures from constrained datasets—mirroring real-world challenges in rule abstraction, cross-linguistic generalization, and constraint satisfaction.

Despite recent advances, current reasoning models exhibit persistent limitations in sparse-data environments and complex symbolic manipulation. Advancing these systems necessitates architectural innovations (e.g., improved verification modules and modality alignment) coupled with rigorous evaluation grounded in cognitively inspired benchmarks, positioning domains like linguistic discovery as critical testbeds for next-generation reasoning capabilities.

\subsection{Reasoning on Linguistic Structures}

Reasoning on linguistic structures presents unique challenges, when compared to other reasoning domains such as math or coding. Unlike purely symbolic systems, understanding human languages requires world knowledge, cultural context, and common sense. For example, the word for ``five'' and ``hand'' is the same in some languages because there are five fingers on a hand. Therefore, reasoning using a symbolic, purely inductive approach makes it difficult to do well in such a linguistically heavy domain.

For the classic Rosetta Stone problems, (that is, given a set of sentences in an unknown language, and their corresponding translations, the agent should infer the underlying rules, such as grammar, meaning of each word, or spelling changes in the unknown language), the inference task is in a sense a more complex variant of the ``infer one form of a word/phrase/sentence to another'' task.

The induction task has long been of interest to linguists (as early as [\citealp{durham-rogers-1969-application}]), as it mirrors what linguists do in a field study. This induction task has been framed in at lease two ways. One perspective treats it as a program synthesis problem, where the goal is to generate a ``program''---a set of formal rules---that transforms inputs to outputs \citep{naik2024largelanguagemodelscode}. This has led to the development of domain-specific languages for expressing such string transformations \citep{vaduguru-etal-2021-sample}. Alternatively, the task can be viewed as constrained text generation, where specialized architectures are designed to model linguistic phenomena \citep{lu-etal-2024-semisupervised}.

A complementary line of research explores augmenting LLMs with explicit linguistic knowledge. Rather than relying solely on induction from examples, this approach provides models with resources like dictionaries, morphological analyzers, or grammar books, mimicking how a human linguist might consult reference materials \citep{zhang-etal-2024-hire}. While the ability to leverage such grammatical descriptions can be systematically evaluated \citep{ICLR2024_52d63f9e}, their utility is task-dependent: for translation, performance gains stem from parallel examples rather than grammatical explanations, which are better suited for targeted linguistic analysis tasks \citep{aycock2025can}. Such nuances call for more research on the intersection of LLMs and linguistics expertise.

\subsection{Relevant Benchmarks from Linguistics Olympiads}

To evaluate the capabilities of LLMs in complex reasoning tasks, researchers have developed various benchmarks. The following are some benchmarks relevant to Linguistics Olympiad problems:

\begin{itemize}
    \item \textbf{LingOly} \citep{lingonly2024}\footnote{Relevant resources for LingOly can be found on GitHub: \url{https://github.com/am-bean/lingOly}.}: This benchmark comprises 1,133 linguistic puzzles from the UK Linguistics Olympiad (UKLO), testing language-agnostic reasoning through diverse formats (e.g., Rosetta, Monolingual) and domains (e.g., morphology, syntax) across 90+ primarily low-resource languages. It excludes image-based puzzles, non-Latin scripts, and open-ended questions to ensure machine-scorability. The evaluation uses an exact-match metric, excluding fuzzy matches and normalizing Unicode variations, to ensure linguistic precision, as the partial credit system used by UKLO human graders cannot be reliably automated. Less strict metrics (i.e., ROUGE and BLEU) are analyzed, but the primary focus remains on context-dependent reasoning.

    \item \textbf{IOLBENCH} \citep{iolbenchbench2025} \footnote{Relevant resources for IOLBENCH can be found on GitHub: \url{https://github.com/Satgoy152/ling_llm}}: This benchmark comprises 90 problems from the IOL archive, ranging from 2003 to 2024. They have digitized and standardized the problems into text or structured representation, including some multimodal components. Each problem is paired with expert-authored solutions to enable fine-grained analysis of reasoning chains. The benchmark is split into text-only and multimodal subsets. The evaluation uses three types of metrics: exact string matching for short answers, automated metrics combined with LLM-based scoring for longer outputs, and manual expert grading for complex reasoning tasks.
    
    \item \textbf{Linguini} \citep{linguini2024}\footnote{Relevant resources for Linguini can be found on GitHub: \url{https://github.com/facebookresearch/linguini}}: This benchmark comprises 160 problems from the IOL, ranging from 2003 to 2023, covering low-resource languages and three core task types: sequence transduction (e.g., script conversion), fill-in-the-blanks (e.g., morphophonological derivation), and number transliteration (e.g., digit-to-text conversion). Problems emphasize skills such as morphosyntactic segmentation, phonological reasoning, and morpheme alignment, often requiring knowledge of linguistic principles such as voicing pairs or coarticulation. The evaluation uses exact match accuracy and the softer chrF metric to assess performance on structured linguistic inference.
        
\end{itemize}

Existing benchmarks for IOL-style tasks have demonstrated the promising capabilities of LLMs in handling complex linguistic reasoning. However, several critical limitations remain that constrain both fine-grained evaluation and meaningful model improvement. 

First, most current evaluations rely predominantly on exact-match accuracy of the final answers, without considering the plausibility, internal consistency, or logical coherence of intermediate reasoning steps. This narrow focus obscures whether models are genuinely applying linguistic principles or merely relying on pattern recognition and heuristic guessing. Such a limitation hampers our ability to diagnose reasoning failures and systematically improve model understanding.

Although step-by-step evaluation approaches, such as Chain-of-Thought (CoT) prompting 
\citep{wei2023chainofthought, wang2023cot, yu2023cot, zhang2022autocot}, have improved interpretability and reasoning transparency in general tasks, they reveal significant shortcomings in the domain of IOL problems. Specifically, these methods often (i) lack rigorous alignment with external linguistic knowledge bases, (ii) fail to capture the reflective, iterative, and self-corrective nature of human linguistic reasoning, and (iii) inadequately represent the hierarchical and multi-layered reasoning structures characteristic of IOL challenges. As a result, existing evaluation paradigms are insufficient for capturing the depth, correctness, and explanatory richness of linguistic problem-solving processes. This highlights the need for more sophisticated evaluation methodologies specifically tailored for linguistic reasoning contexts.

Second, many IOL benchmarks lack authoritative gold-standard solutions with detailed linguistic analyses and rule-based derivations. The absence of such structured explanations limits the capacity to evaluate model outputs beyond surface-level correctness and hinders the development of explainability-driven evaluation protocols. Third, typological and linguistic metadata about the languages involved in IOL problems are frequently omitted. This omission restricts systematic investigations into LLM generalization capabilities across diverse linguistic families, structures, and typological categories, limiting insights into model strengths and weaknesses in cross-linguistic reasoning.

To address these critical gaps, we propose a novel benchmark specifically designed for IOL tasks, which incorporates comprehensive typological metadata, and explainability-oriented evaluation protocols. By centering evaluations on reasoning quality rather than solely on final answer accuracy, this benchmark aims to provide more diagnostic insights and foster targeted improvements in future LLM architectures.

\section{Linguistics Olympiad Benchmark for Structured Evaluation on Reasoning}
\label{sec:LOBSTER}


\subsection{Motivation and Design Goals}

\begin{figure}
    \centering
    \includegraphics[width=0.9\textwidth]{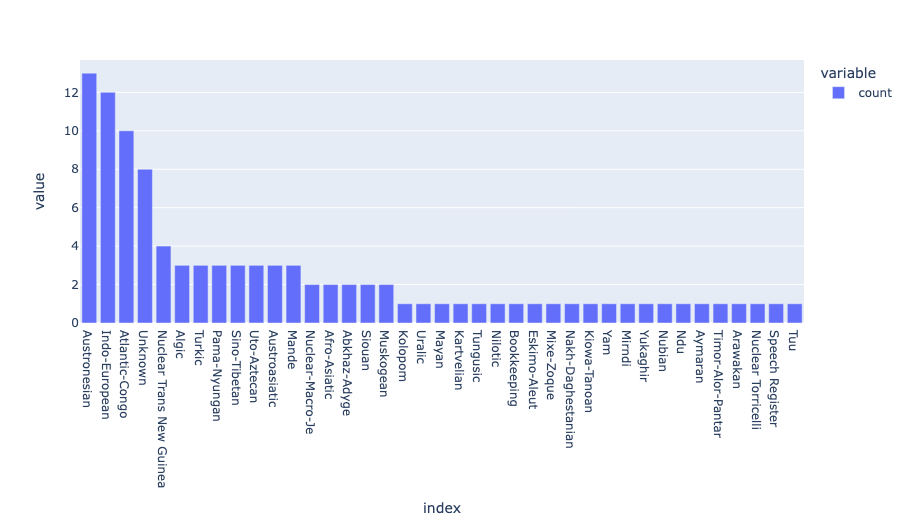}
    \caption{IOL Problem Distribution by Language Family}
    \label{fig:iol-problem-family}
\end{figure}

Regarding the distribution of IOL problems by language family, the most common language families are Austronesian, Indo-European, and Atlantic-Congo (Figure~\ref{fig:iol-problem-family}). However, despite the apparent linguistic diversity involved in the IOL problems, existing LLM benchmarks rarely account for such distributional factors. There remains a gap in understanding how models perform across different language families and typological features.

Although recent benchmarks have brought valuable attention to LLM performance on linguistic puzzles, they suffer from critical limitations: transcription errors (e.g., Problem 2008-1 answers leaked and Problem 2005-4 tone markers mislabeled in IOLBENCH), unverified task categorization, and inadequate handling of non-textual data (e.g., Problem 2016-2 image-to-text conversion artifacts). More fundamentally, such benchmarks primarily offer black-box evaluations focused on final output accuracy via exact-match metrics, while overlooking the model’s reasoning process, logical thinking, and adherence to linguistic rules—aspects typically relegated to costly human evaluation. To address these gaps, we propose LingBench++, a linguistically-informed benchmark designed for stepwise, interpretable evaluation of LLMs in solving IOL-style tasks, combining rigorous problem curation with transparent scoring of intermediate reasoning steps.

\textbf{LingBench++} is built on a curated selection of past IOL problems. Unlike prior datasets, it includes enriched metadata that allows for deeper linguistic diagnostics and reasoning trace comparison. Our benchmark is intended to support the following:
\begin{itemize}
\item Accurate transcription of contents of IOL problems.
\item Evaluation of reasoning steps (via gold-standard explanation traces).
\item Typologically grounded performance analysis.
\item Assessment of models' cross-cultural and cross-linguistic inference abilities.
\item Comparison between single-shot vs. agentic reasoning approaches.
\end{itemize}

\subsection{Data Construction}
\label{sec:benchmark-construction}
Our benchmark consists of 96 problems (225 sub-problems) sourced from the IOL archive (2003–2024). For kinship problems\footnote{Kinship problems focus on understanding how different languages and cultures describe family relationships and naming systems.} involving family trees, we convert the graphical representations into textual relationship descriptions (see Appendix \ref{appendix:kinship-example}). We exclude problems that fully rely on image-based information or untranscribable symbols.

Since most IOL problems provide only the final solutions along with some grammatical rules, without including detailed reasoning steps, we use Gemini-2.5-Pro to generate structured step-by-step solutions as gold-standard references in LingBench++. The LLM is prompted to act as a linguistics expert, producing logical deductions, linguistic rules, and problem-solving strategies that lead to the official solutions (see Appendix~\ref{appendix:prompt-template} for the prompt template). To ensure reliability, seven human experts and three IOL contestants manually verify and refine these reasoning chains, resolving any inconsistencies to ensure alignment with the official IOL solutions. In summary, for each IOL problem in our benchmark, we include the transcribed problem text, the official solution, and the expert-verified, refined LLM-generated reasoning.

\subsection{Typological Annotation} 

In addition, each problem within LingBench++ is annotated along multiple linguistic dimensions to facilitate a structured analysis of model performance. The current typological and problem-oriented schema is an adaptation of the UKLO classification framework\footnote{\url{https://www.uklo.org/technical-information/}}. This annotation process is carried out by seven linguistic experts. We annotate three categories for each problem: Subject, Type, and Theme; the respective tags are detailed below, while the descriptions of each tag are shown in Appendix~\ref{appendix:uklo-classification}. Also, the corresponding language, language family, Glottocode, and the number of speakers of the language are also manually recorded for each problem. Table~\ref{table:annot-example} shows an example of annotations for one problem.

\begin{table*}[ht]
\centering
\begin{threeparttable}
\begin{tabular}{>{\bfseries}l p{0.8\linewidth}}
\toprule
\textbf{Category} & \textbf{Tag} \\
\midrule
Subject & Compounding, Morphology, Numbers, Phonology and Phonetics, Semantics, Syntax, Writing System \\
Type & Rosetta, Match-up, Monolingual, Pattern, Computational, Text \\
Theme & Classical, Comparative, Encrypted, Kinship, Maps, Mystery, MFL\tnote{1}, Senses and Feelings, Stories, Poetry, No Theme \\
\bottomrule
\end{tabular}
\begin{tablenotes}
\footnotesize
\item[1] MFL: These questions involve languages commonly taught in secondary school MFL departments, or those closely related (e.g., Romance and Germanic languages).
\end{tablenotes}
\end{threeparttable}
\caption{Typological Annotation Category}
\label{table:annot-category}
\end{table*}

\begin{table}[h!]
\small
\centering
\renewcommand{\arraystretch}{1.2}
\begin{tabularx}{\textwidth}{lllllllll}
\toprule
\textbf{Sub-problems} & \textbf{Subject} & \textbf{Type} & \textbf{Language} & \textbf{Speakers} & \textbf{glottocode} & \textbf{Language Family} \\
\midrule
2 & Numbers & Pattern & Egyptian Arabic & 68,000,000 & egyp1253 & Semitic \\
\bottomrule
\end{tabularx}
\caption{Example of Typological Annotation: Problem 2 in 2003}
\label{table:annot-example}
\end{table}

\subsection{Preliminary Data Analysis}
\begin{figure}[htb!]
    \centering
        \includegraphics[width=0.9\textwidth]{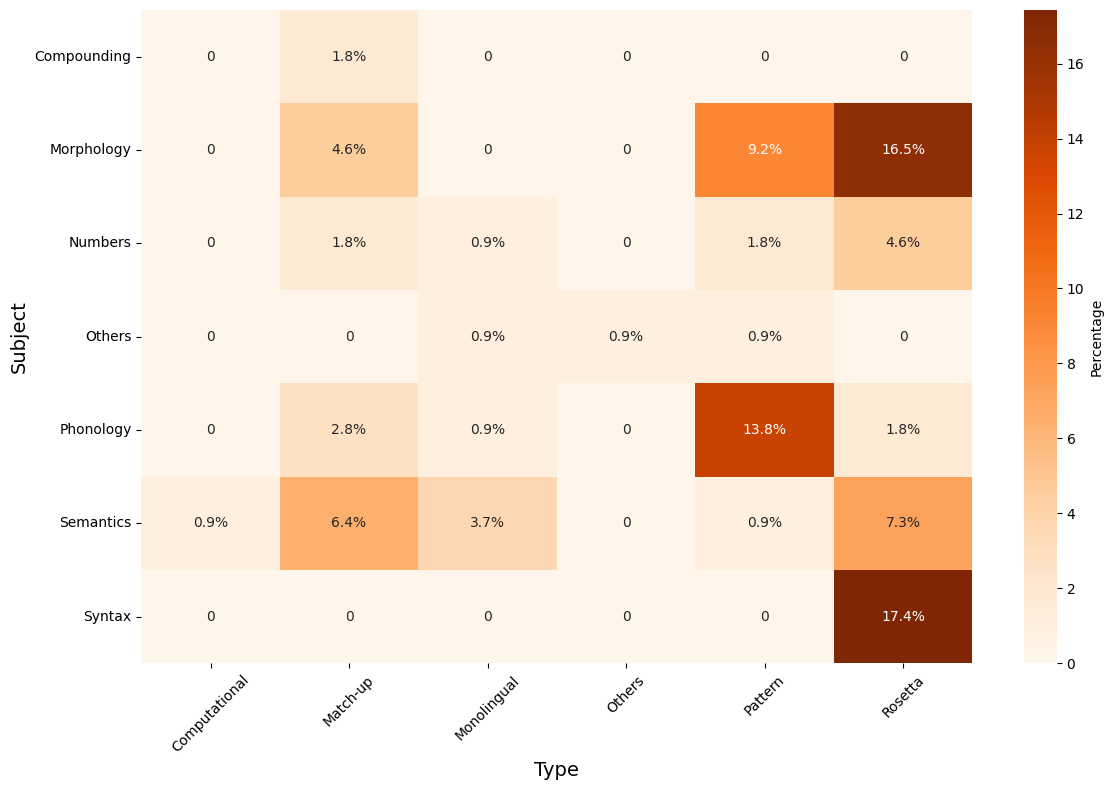}
        \caption{Subject vs Type Distribution}
        \label{fig:subject-type}
\end{figure}
\begin{figure}[htb!]
    \centering
        \includegraphics[width=0.9\textwidth]{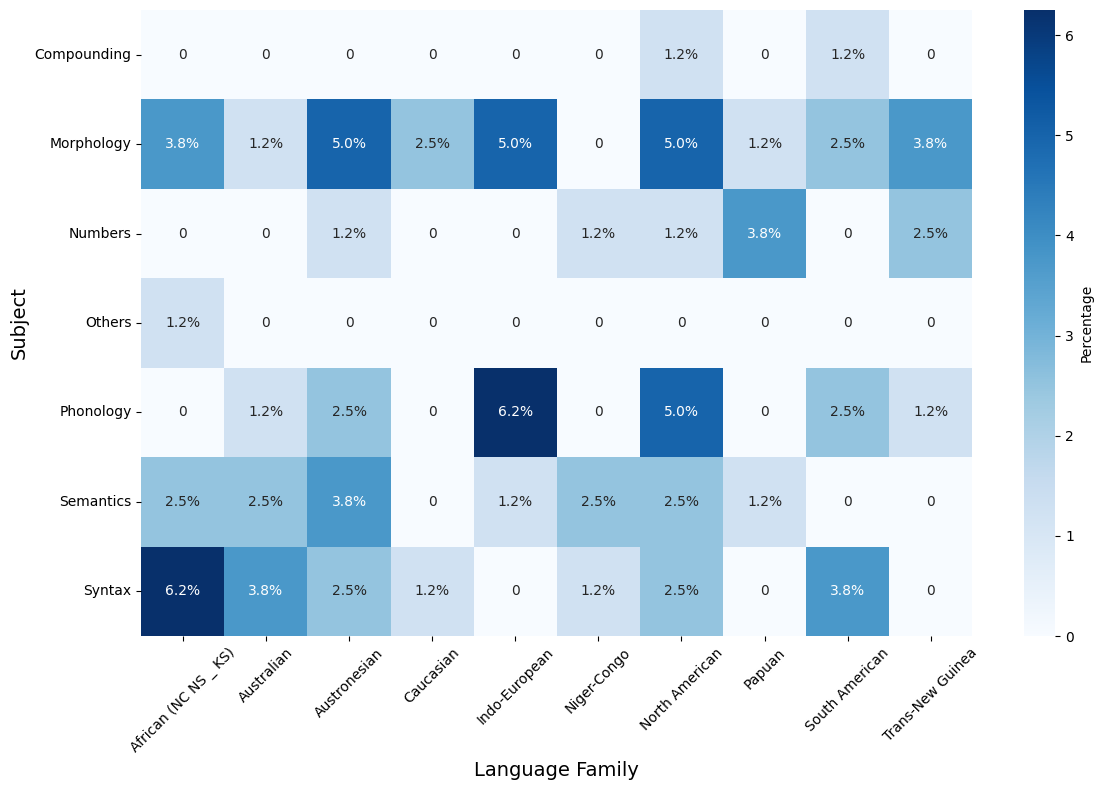}
        \caption{Subject vs Top 10 Language Family Distribution}
        \label{fig:subject-langfamily}
\end{figure}
\begin{figure}[htb!]
    \centering
        \includegraphics[width=0.9\textwidth]{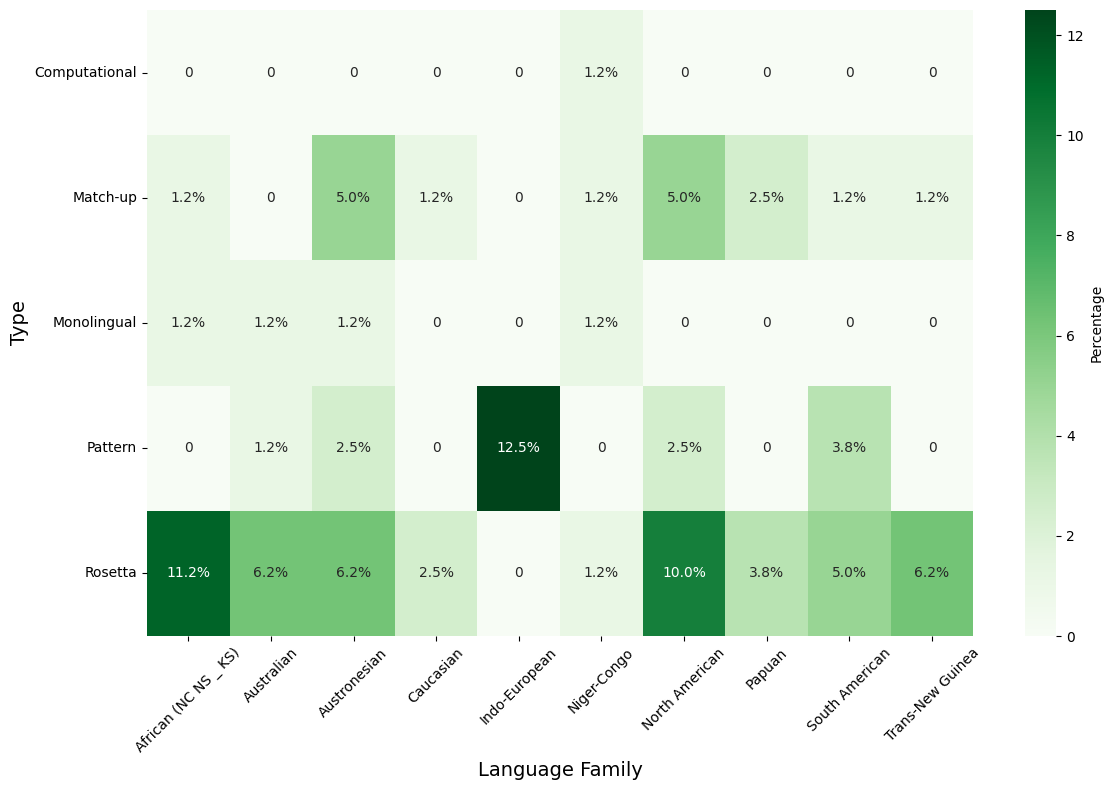}
        \caption{Type vs Top 10 Language Family Distribution}
        \label{fig:type-langfamily}
\end{figure}

The distribution charts of each typological category in our benchmark are shown in Appendix~\ref{appendix:preliminary-analysis}. Key findings include:

\vspace{0.5em}
\noindent \textbf{Subject and Type Distribution}: As shown in Figure~\ref{fig:subject-type}, the data suggests that Syntax and Morphology are the most prominent subjects in IOL problems, with Rosetta type problems being heavily focused in these areas (i.e., 17.4\% and 16.5\%). Semantics are distributed across multiple problem types (0.9\%, 6.4\%, 3.7\%, 0.9\%, 7.3\%) compared to others. Overall, the uneven distribution implies that certain problem types are strongly associated with particular subjects (e.g., Phonology has a spike (13.8\%) in Pattern type problems), while others are more diffuse.

\vspace{0.5em}
\noindent  \textbf{Subject and Language Family Distribution}:  North American languages have the highest number of problems (14), followed by Austronesian (11), Indo-European (10), and African (10). As shown in Figure~\ref{fig:subject-langfamily}, Syntax is the most widely represented subject, appearing in 7 out of the top 10 language families, with the highest concentration (6.2\%) in African. Morphology is the second most frequent, appearing in 9 out of the top 10 families, with multiple mid-range values (2.5\%–5.0\%). While Phonology stands out in Indo-European and North American, Semantics is more broadly distributed, with Austronesian, African, Australian, and Niger-Congo all having moderate percentages (around 2.5\%). In summary, Syntax, Morphology, and Phonology dominate the subject distribution, with North American, Austronesian, Indo-European, and African languages showing the richest variety of subjects. More details are shown in Figures (a) and (d) in Appendix~\ref{appendix:preliminary-analysis}.

\vspace{0.5em}
\noindent \textbf{Type and Language Family Distribution}: As shown in Figure~\ref{fig:type-langfamily}, \textit{Match-up} problems are more common in Austronesian and North American language families. \textit{Pattern} problems are particularly prevalent in Indo-European languages. \textit{Rosetta} problems are the most common overall (44 problems), appearing across various language families, with especially high occurrences in African and North American languages. More details are shown in Figure \ref{fig:iol-problem-stats} (b) and (d) in Appendix \ref{appendix:preliminary-analysis}.

These findings reinforce the relevance of typological and reasoning-aware annotations. They also highlight the inadequacy of answer-only metrics in capturing the richness of linguistic cognition demanded by IOL tasks.

\subsection{Evaluation Protocol and Metrics}

Existing benchmarks for IOL-style tasks often lack a unified framework that can balance precision with flexibility. Current approaches tend to rely on strict exact-match accuracy, which fails to award partial credit for complex problems. Moreover, the representation of IOL answers is rather complex. Some accept multiple possible answers, while some require an explanation of the rules. To address these limitations, we adjust the evaluation for the final solution generated by the model, considering the rules provided in official solutions. Also, we provide a protocol that may be applied when evaluating future model reasoning.

\subsubsection{Evaluation of the Final Solution}
\label{sec:eval-final-solution}
First, we assess the model-generated final solution based on two distinct components: the \textbf{answer} and the \textbf{explanation of rules}.

The \textbf{answer} refers to the specific outputs the problem explicitly requests, such as providing translations, filling in tables, or matching items. The main challenge in grading answers is that different subproblems require different types of answers, ranging from matching, translation (sometimes there are more than one possible translations) to explaining why a sentence is incorrect\footnote{Note: This should not be confused with the ``explanation'' part below.}. By default, exact match is applied to grade the answers. The reference answer for each problem has been converted into a structured JSON format, and the model is required to output accordingly. However, for subproblems where an exact match is unsatisfactory, additional annotations are tagged, including:
\begin{itemize}
\item Some subproblems may be tagged \texttt{<fuzzy>}, meaning the answer should be evaluated based on meaning rather than strictly graded verbatim (e.g., explanation problems). In these cases, another metric of the user's choice should be applied (e.g., BLEU, sentence embedding, or an LLM judge).
\item To handle the multiple-answer case, the reference answer is a list with the tag \texttt{<select\%d, \%d, \%d>}. The tag contains 3 arguments: the length of the correct answer, the minimum number of the model's output, and the maximum number of the model's output. For example, if there's a translation task with description ``Give the all the possible translation of the sentence'' and the ground truth has 2 possible translations, the tag will be \texttt{<select2, 1, inf>}. The \texttt{1, inf} indicates that, when prompted to solve the problem, the model would not know the number of correct answers in advance, and thus could output one to an arbitrary number of answers.
\end{itemize}




On the other hand, the \textbf{explanation} requires the model to write down the linguistic rules it inferred from the problem data. Since official IOL grading rubrics are not publicly available, evaluating the quality of these free-text explanations poses a significant challenge unaddressed by past works. We address it with a two-stage procedure:

\begin{itemize}
\item Rule Decomposition: For each problem, we manually decomposed the official solution into a discrete set of key linguistic rules, creating a gold-standard ``rule checklist.''
\item Checklist Grading: We then employ an LLM in our grading process. The grader is prompted to compare the model's generated explanation against our rule checklist and determine the number of gold-standard rules that were correctly described.
\end{itemize}

This approach enables a stable, fine-grained, and quantitative assessment of the explanation's quality. The total score for the final solution is a weighted combination of the scores of ``answer'' and ``explanation of rules.'' By default, we assign equal weight (50/50) to each component, with points distributed evenly across all sub-problems (for the Answer) and all identified rules (for the Explanation). This weighting is fully customizable to suit different analytical priorities.

It is important to note that this section evaluates the outcome of the reasoning process. The step-by-step process of discovering these rules is assessed separately as part of the reasoning evaluation.

\subsubsection{Statistical Analysis of the Final Solution}

\begin{figure*}[htb!]
    \centering
        \includegraphics[width=0.9\textwidth]{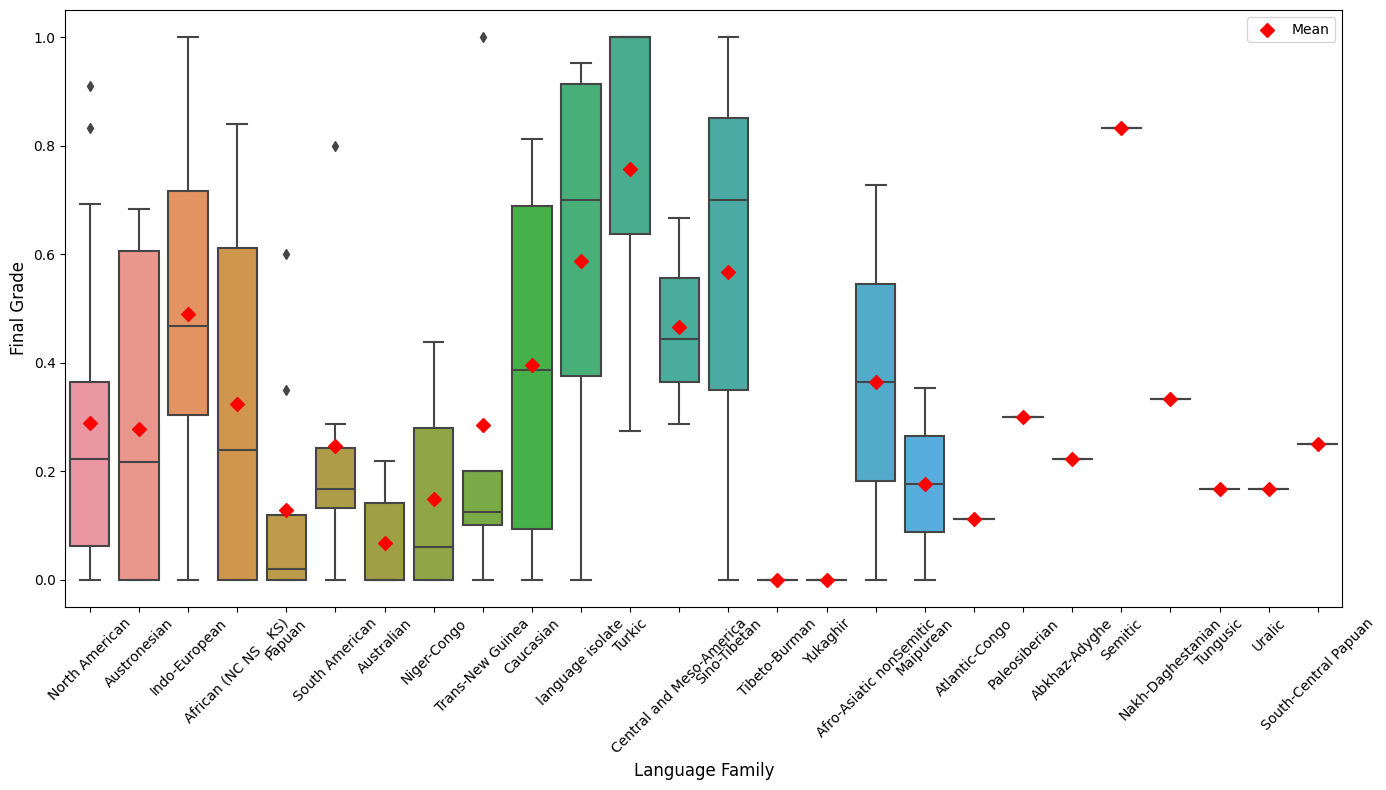}
        \caption{Distribution of Final Grades by Language Family}
        \label{fig:dist-final-grades-lang-family}
\end{figure*}

Based on the evaluation methods for the final solutions in Section~\ref{sec:eval-final-solution}, we apply our multi-agent framework to the 96 problems in LingBench++ and conduct a statistical analysis based on the typological tags assigned to each problem. 

Figure~\ref{fig:dist-final-grades-lang-family} displays the distribution of final scores across various language families. The model demonstrates notable performance variations among language families. It achieves the highest scores on Turkic (mean = 0.76) and Semitic (0.83) problems, often reaching perfect scores, while also performing well on language isolates (mean = 0.59) and Sino-Tibetan (0.57) problems. Performance is moderate for Indo-European (0.49), Central and Meso-American (0.47), and Caucasian (0.40) families, though with considerable variability (e.g., Indo-European std = 0.34). In contrast, the model struggles significantly with Australian (mean = 0.07), Papuan (0.13), Niger-Congo (0.15), and Tibeto-Burman (0.00) problems, consistently yielding low scores. High variability is particularly evident in Trans-New Guinea (std = 0.41) and Afro-Asiatic non-Semitic (std = 0.51) families, reflecting unstable reasoning patterns. Additionally, results from families with very few samples (e.g., Atlantic-Congo, Paleosiberian) should be interpreted cautiously due to limited data reliability.

\begin{figure*}[htb!]
    \centering
        \includegraphics[width=0.9\textwidth]{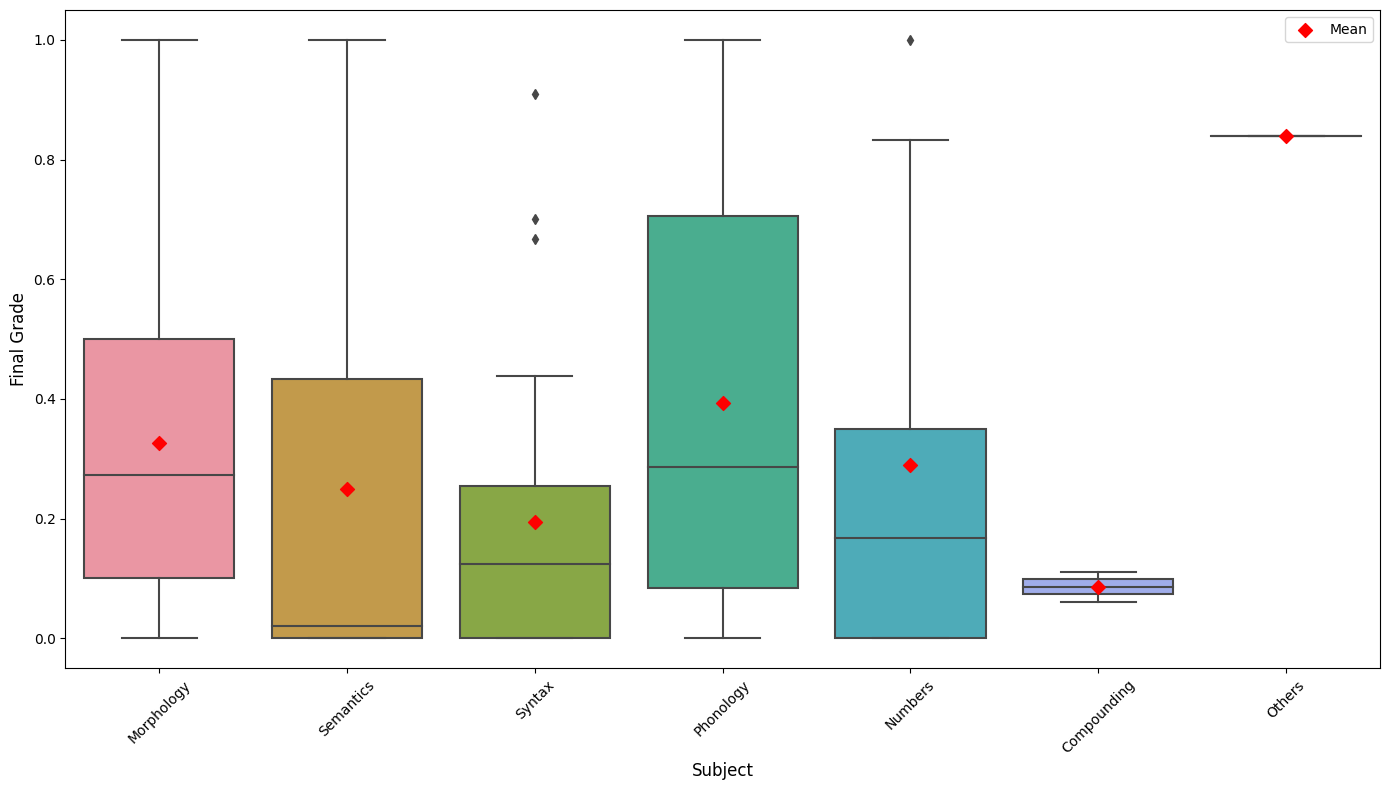}
        \caption{Distribution of Final Grades by Subject}
        \label{fig:dist-final-grades-subject}
\end{figure*}

Figure~\ref{fig:dist-final-grades-subject} presents the distribution of final grades across different linguistic subjects. The model exhibits notable performance variations, with the strongest results observed in \textit{Phonology} (mean accuracy = 0.39), though substantial variability ($\sigma$ = 0.35) indicates inconsistent performance. \textit{Morphology} shows moderate performance (0.33), followed by \textit{Numbers}-related tasks (0.29), suggesting limited but detectable reasoning capabilities. The model performs poorly in \textit{Semantics} (0.25) despite the subject's theoretical proximity to phonology, with particularly high variability (std 0.33) reflecting unpredictable outcomes. Performance is weakest in \textit{Syntax} (0.19) and \textit{Compounding} (0.09), where the model consistently struggles, with compounding problems showing minimal variation (std 0.035). While the \textit{Others} category displays exceptional accuracy (0.84), the single data point prevents meaningful generalization. Collectively, these results suggest the model handles sound-based patterns best but falters significantly in structural and meaning-based linguistic tasks.

\begin{figure*}[htb!]
    \centering
        \includegraphics[width=0.9\textwidth]{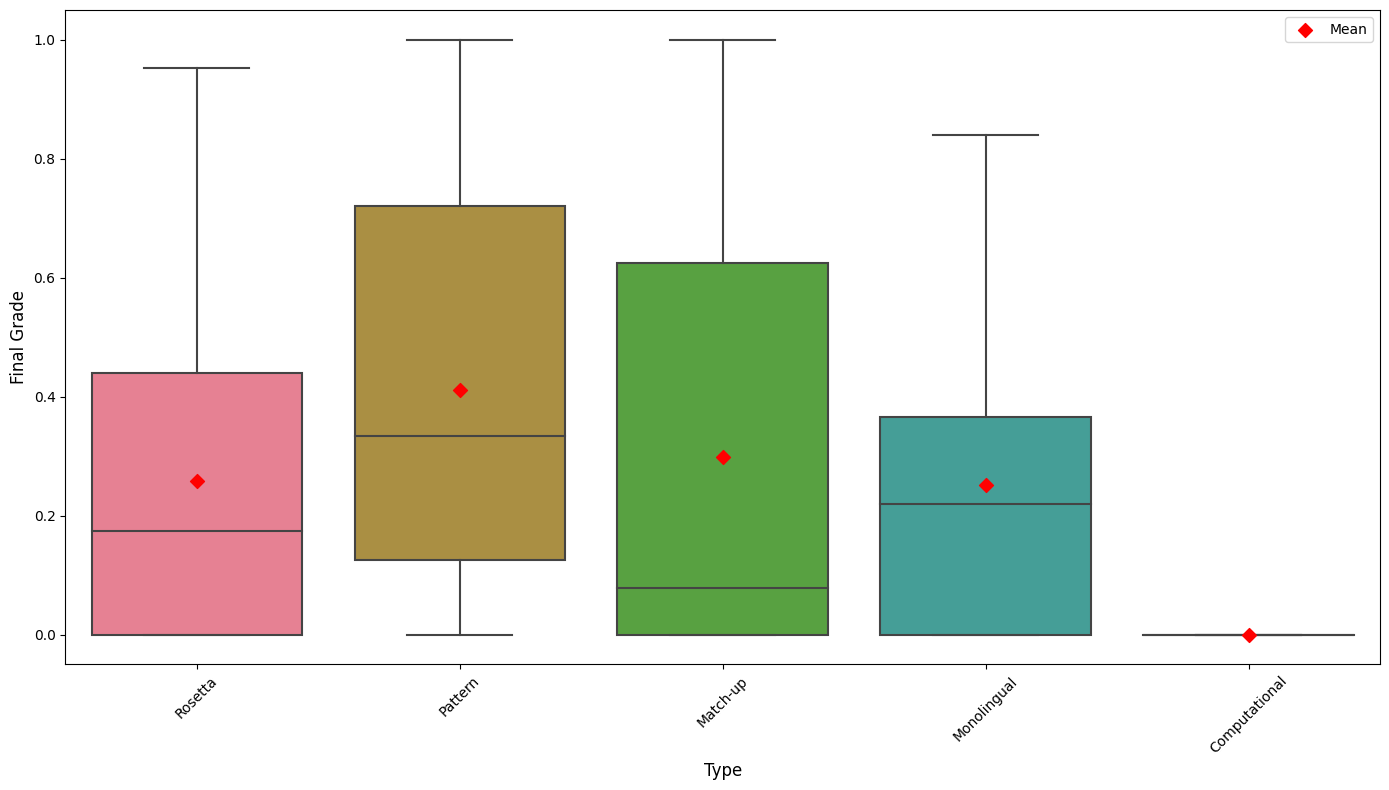}
        \caption{Distribution of Final Grades by Type}
        \label{fig:dist-final-grades-type}
\end{figure*}

For problem types (Figure~\ref{fig:dist-final-grades-type}), the model performs worst on \textit{Rosetta} (mean = 0.259, median = 0.174), indicating significant difficulty. \textit{Pattern} problems achieve the highest scores (mean = 0.411, median = 0.333) but with substantial variability ($\sigma$ = 0.347). \textit{Match-up} shows moderate performance (mean = 0.299) but extreme inconsistency (median = 0.078, $\sigma$ = 0.371). \textit{Monolingual} tasks (mean = 0.252, median = 0.219) have limited generalizability ($n$ = 11), while the single \textit{Computational} instance (score = 0.0) is statistically insignificant. Overall, the model struggles most with \textit{Rosetta}, performs best (though inconsistently) on \textit{Pattern}, and shows instability in \textit{Match-up}.



The typological analysis shows the model performs well on Sino-Tibetan, Turkic, and language isolate problems, especially in phonology and semantics. However, it struggles with Australian, Papuan, and Atlantic-Congo problems, as well as syntax and compounding tasks. By problem type, it performs worst on Rosetta problems and better, though inconsistently, on Pattern and Match-up types. Overall, the model shows strong performance in certain areas but inconsistent reasoning across languages, subjects, and problem types.

\subsubsection{Evaluation of Reasoning Steps}

In addition to evaluating the correctness of final answers, we propose a Check-of-Thought protocol, with an emphasis on assessing the quality of the reasoning process. This is motivated by two key considerations: first, a model may produce incorrect final answers while still demonstrating plausible reasoning; second, it may arrive at correct answers through guessing or disorganized reasoning. 

The proposed scoring dimensions and corresponding goals are listed below; (i), (ii), and (iii) align the target model reasoning with the gold reasoning reference from our benchmark, while (iv) and (v) consider only the target model reasoning. The prompt for the judges are shown in the Appendix \ref{appendix:eval-prompt-metric-example}. These aspects of evaluation ensures that the target model's reasoning process are systematically examined. 

\begin{itemize}
    \item (i) Information Extraction \& Structuring
    \begin{itemize}
        \item \textbf{Stepwise Logical Validity Score (SLVS)}: Measures whether each reasoning step is logically valid and aligned with the gold reasoning reference (GRR).
        \item \textbf{Information Structuring Completeness (ISC)}: Measures the completeness of the extracted key information and its structure compared to the GRR.
    \end{itemize}
    
    \item (ii) Hypothesis Generation \& Rule Induction
    \begin{itemize}
        \item \textbf{Hypothesis Generation Adequacy (HGA)}: Measures plausibility and sufficiency of generated hypotheses relative to the GRR.
        \item \textbf{Rule Induction Coverage (RIC)}: Measures how well the induced rules cover the core rules identified in the GRR.
        \item \textbf{Inference Justification Coverage (IJC)}: Measures the degree to which reasoning steps include explicit justifications matching the GRR.
    \end{itemize}
    
    \item (iii) Completeness \& Coverage
    \begin{itemize}
        \item \textbf{Chain-of-Thought Continuity Score (CCS)}: Measures continuity and coherence of the reasoning chain relative to the GRR.
        \item \textbf{Subtask Coverage Rate (SCR)}: Measures coverage of all subtasks or sub-questions included in the GRR.
    \end{itemize}
    
    \item (iv) Logical Deduction \& Rule Application (Model Internal)
    \begin{itemize}
        \item \textbf{Stepwise Logical Validity Score (SLVS)}: Measures internal logical consistency and validity of reasoning steps without referencing the GRR.
        \item \textbf{Application Consistency Rate (ACR)}: Measures consistency in rule application within the model's reasoning process.
    \end{itemize}
    
    \item (v) Contradiction Handling \& Self-Revision (Model Internal)
    \begin{itemize}
        \item \textbf{Contradiction Detection Ability (CDA)}: Measures ability to detect contradictions within the model's own reasoning.
        \item \textbf{Error Traceability Score (ETS)}: Measures ability to trace and correct errors internally.
    \end{itemize}
\end{itemize}

\begin{table}[htbp!]
\centering
\begin{tabular}{>{\raggedright}p{3cm}llp{8cm}}
\toprule
\textbf{Dimension} & \textbf{Metric} & \textbf{Score} & \textbf{Justification} \\
\midrule
\textbf{(i) Information Extraction \& Structuring} & SLVS & 2 & Some steps are logically valid (e.g., identifying the ``X threatens Y'' structure), but key insights (e.g., desirability dichotomy) are missing. \\
 & ISC & 2 & Extracted structure is incomplete; misses critical grouping of headlines by desirability of objects. \\
\midrule
\textbf{(ii) Hypothesis Generation \& Rule Induction} & HGA & 1 & Hypotheses (e.g., ``reversal of causality'') are implausible and misaligned with GRR. \\
 & RIC & 1 & Induced rules (e.g., ``reverse the relationship'') do not cover the GRR's core rules (desirability-based ambiguity). \\
 & IJC & 2 & Justifications (e.g., ``tautological relationships'') are weakly supported and deviate from the GRR. \\
\midrule
\textbf{(iii) Completeness \& Coverage} & CCS & 2 & Reasoning chain lacks continuity (e.g., skips desirability analysis entirely). \\
 & SCR & 1 & Fails to cover subtasks (e.g., no classification of all 11 headlines). \\
\midrule
\textbf{(iv) Logical Deduction \& Rule Application} & SLVS & 1 & Internal logic is inconsistent (e.g., arbitrary reversal of relationships). \\
 & ACR & 1 & Rule application is ad hoc (e.g., inconsistently reverses only 3 titles). \\
\midrule
\textbf{(v) Contradiction Handling \& Self-Revision} & CDA & 1 & No detection of contradictions (e.g., ignores ambiguity in titles 7/9). \\
 & ETS & 1 & No evidence of error tracing or correction (e.g., persists with flawed reversal logic). \\
\bottomrule
\end{tabular}
\caption{Example of Evaluation Results for Model Reasoning. GRR stands for the golden reasoning reference.}
\label{table:eval-reasoning-result}
\end{table}

However, due to the substantial manual effort required, we introduce the possible prompt and results of evaluating a single problem rather than performing a large-scale application of these metrics on model reasoning. Specifically, the Check-of-Thought protocol involves fine-grained human-in-the-loop judgments across multiple reasoning dimensions. Implementing this evaluation at scale would require significant annotator time and cost, especially for linguistically diverse IOL problems with complex reasoning chains. Therefore, in this sub-section, we mainly introduce the evaluation framework and scoring guidelines, leaving large-scale application and automation of reasoning evaluation as future work.

We apply our refined reasoning process from the benchmark (Section~\ref{sec:benchmark-construction}) and use another LLM as a judge to assess the reasoning of the baseline model (Section~\ref{sec:agent-result-and-analysis}) by comparing it to the gold reasoning reference, using Problem 2 from 2004 for example. Specifically, we require the judge LLM to score the reasoning based on the prompt, which includes the description and scoring criteria for five dimensions, the gold reasoning reference, and the baseline model's reasoning (see Appendix~\ref{appendix:eval-prompt-metric-example}). 

The evaluation results presented in Table \ref{table:eval-reasoning-result} provide a systematic, granular, and actionable framework for assessing model reasoning and assigning scores with clear justifications. This enables identification of weaknesses (e.g., low HGA revealing flawed hypotheses or low CDA exposing missed contradictions), offers targeted feedback for improvement in model reasoning (e.g., addressing incomplete subtask coverage flagged by SCR), and ensures transparency by linking scores to explicit criteria. By evaluating both alignment with a gold reasoning reference (GRR) and internal validity (e.g., self-consistency via ACR), the metrics deliver insights while standardizing comparisons across models. Ultimately, this method transforms subjective reasoning evaluation into an interpretable, reproducible tool.


\section{Utilizing Reference Grammar for Confirmation and Verification}
\label{section:utilizing_ref_grammar}

\subsection{Limitations of Pre-Existing LLM Knowledge}

While LLMs gain vast amounts of knowledge during pre-training, the knowledge that can be acquired is limited to what resources are available. \citet{joshi-etal-2020-state} categorize languages according to the number of labeled and unlabeled resources, with Class \texttt{0} having ``exceptionally limited resources'' and Class \texttt{5} being ``the quintessential rich-resource languages, reaping benefit from each state-of-the-art NLP breakthrough.''\footnote{The list of languages and their classes can be found here: \url{https://microsoft.github.io/linguisticdiversity/assets/lang2tax.txt}} 
They indicate that almost 90\% of languages are low-resource with virtually no labeled or unlabeled data (i.e., category 0), such as Warlpiri, an Australian Aboriginal language. 
This means that LLMs are actually exposed to only a small proportion of the world's languages. Furthermore, the languages that an LLM sees during training are disproportionately from rich resource languages, such as English or Spanish, that have a dominant presence online.

\cite{joshi-etal-2020-state} also examine this issue from a linguistic typology perspective, which involves classifying languages based on their structure and semantics. At the time of publication, they calculated that across languages with limited or no data resources (i.e., categories 0-2), there are 549 out of 1139 unique categories across 192 typological features that are not found in the higher-resourced categories of 3, 4, and 5.
Languages with limited typological representation in training data also adversely affect their performance across common NLP tasks, such as similarity search, as demonstrated by the authors. This limited representation hinders cross-lingual model transfer as low-resource languages cannot benefit from higher-resourced languages that share typological features. 
\citet{pires-etal-2019-multilingual} have demonstrated that models are capable of performing model transfer across typologically similar languages.

\subsection{Evaluating SOTA Model on Multilingual Translation}
To understand how well current SOTA models perform in a variety of languages, we test \texttt{gemini-2.5-flash} on a subset of the FLORES-200 benchmark dataset,\footnote{Dataset is available here: \url{https://huggingface.co/datasets/facebook/flores}.} which was developed as a result of the efforts of the No Language Left Behind (NLLB) project. The goal of this project is to prioritize the needs of low-resource language communities and to train models that narrow the performance gap between low- and high-resource languages \citep{nllbteam2022languageleftbehindscaling}. FLORES-200 is a many-to-many multilingual dataset aimed at evaluating translation quality through more than 40,000 translation directions. It consists of 3001 sentences sampled from English-language Wikimedia projects that were professionally translated into 200+ languages (including different scripts).

\paragraph{Dataset preparation and experimental design.} We combine the \texttt{dev} and \texttt{devtest} splits for a total of 2009 sentences that are available in \textbf{204} languages. We then use the ISO 639-3 language code and the ISO 15924 script code to identify the Glottocode and the script used for each language, respectively. For example, the column name for Bashkir translations written in Cyrillic is \texttt{sentence\_bak\_Cyrl}.
To align the dataset with the Glottolog taxonomy, we mapped all language identifiers to their corresponding Glottolog codes. We noted that five ISO 639-3 codes from the dataset (i.e., \texttt{srd}, \texttt{est}, \texttt{kon}, \texttt{zho}, \texttt{grn}) were not directly linked to a Glottolog entry. We identified suitable entries manually. How we mapped these languages can be found in Table~\ref{tab:iso-glottolog-mapping}. In total, we have 204 languages and script combinations.\footnote{196 unique languages while Acehnese, Minangkabau, Banjar, Central Kanuri, Tamasheq, Standard Arabic, Kashmiri, and Mandarin each have two scripts.}
Next, we take the first \textbf{10} English sentences and their translations for a total of \textbf{2030} English-to-Target Language pairs. 

We evaluate \texttt{gemini-2.5-flash} with \texttt{temperature=0.1} and \texttt{thinking budget=0} by translating from two directions: \emph{English-to-Target} ($E \rightarrow T$) and \emph{Target-to-English} ($T \rightarrow E$). We use the following $E \rightarrow T$ prompt when eliciting a response from the model: 
\begin{quote}
\texttt{Translate the following sentence from English to \{target\_lang\} using the \{script\} script:} \\
\texttt{Input: \{input\_sentence\}}
\end{quote}

We use the following $T \rightarrow E$ prompt:
\begin{quote}
\texttt{Translate the following sentence \{target\_lang\} to English:} \\
\texttt{Input: \{input\_sentence\}}
\end{quote}

\begin{table}[htbp]
\begin{tabular}{llcrrr}
\toprule
\multicolumn{1}{m{3.5cm}}{\centering\textbf{Language}} &
\multicolumn{1}{m{2cm}}{\centering\textbf{Glottocode}} &
\multicolumn{1}{m{1cm}}{\centering\textbf{Class}} &
\multicolumn{1}{m{2cm}}{\centering\textbf{Missing \\ ($E \rightarrow T$)}} &
\multicolumn{1}{m{2cm}}{\centering\textbf{Missing \\ ($T \rightarrow E$)}} &
\multicolumn{1}{m{1.5cm}}{\centering\textbf{Total \\ Missing}} \\
\midrule
Tamasheq                  & tama1365 & 0   & 7 & 1 & 8 \\
Nuer                      & nuer1246 & 0   & 6 & 2 & 8 \\
Kabiyé                    & kabi1261 & 0   & 7 & 0 & 7 \\
Southwestern Dinka        & sout2832 & --- & 6 & 1 & 7 \\
Central Kanuri            & cent2050 & 0   & 4 & 2 & 6 \\
Fon                       & fonn1241 & 0   & 5 & 0 & 5 \\
Chokwe                    & chok1245 & --- & 2 & 1 & 3 \\
Umbundu                   & umbu1257 & 0   & 3 & 0 & 3 \\
Kamba (Kenya)             & kamb1297 & 0   & 2 & 0 & 2 \\
Sango                     & sang1328 & 1   & 2 & 0 & 2 \\
South-Central Koongo      & koon1244 & 1   & 2 & 0 & 2 \\
Kimbundu                  & kimb1241 & 0   & 2 & 0 & 2 \\
Bambara                   & bamb1269 & 1   & 2 & 0 & 2 \\
Dyula                     & dyul1238 & 0   & 2 & 0 & 2 \\
Mossi                     & moss1236 & 0   & 4 & 0 & 4 \\
Southern Jinghpaw         & kach1280 & 0   & 4 & 0 & 4 \\
Shan                      & shan1277 & 0   & 4 & 0 & 4 \\
Acehnese                  & achi1257 & 1   & 1 & 0 & 1 \\
Ewe                       & ewee1241 & 1   & 1 & 0 & 1 \\
Dzongkha                  & dzon1239 & 1   & 1 & 0 & 1 \\
Central Aymara            & cent2142 & --- & 1 & 0 & 1 \\
Ayacucho Quechua          & ayac1239 & --- & 1 & 0 & 1 \\
Luba-Lulua                & luba1249 & 0   & 1 & 0 & 1 \\
Kabyle                    & kaby1243 & 1   & 1 & 0 & 1 \\
Guarani                   & east2555 & 1   & 1 & 0 & 1 \\
Wolof                     & nucl1347 & 2   & 1 & 0 & 1 \\
\midrule
\textbf{Grand Total} & & & \textbf{73} & \textbf{7} & \textbf{80} \\
\bottomrule
\end{tabular}
\caption{Counts of missing LLM Outputs by language and direction. \emph{Class} refers to the taxonomy introduced in \citet{joshi-etal-2020-state} in which \texttt{0} indicates extremely limited resources and \texttt{5} indicates an abundance of resources. ``--'' means that the language was not found in the taxonomy.}
\label{tab:missing_outputs_class}
\end{table}

With the LLM translating in two directions, we obtain 3800 responses; however, 80 responses are empty with the majority of them originating from the $E \rightarrow T$ task. We will first examine these failures.

\paragraph{The LLM often fails to output any text for low resource languages.} From the results in  Table~\ref{tab:missing_outputs_class} we can see that the data strongly suggests that the model's failure to generate output is directly linked to data resource scarcity. The \emph{Class} column refers to the taxonomy introduced in \citet{joshi-etal-2020-state} where Class \texttt{0} languages have a dearth of resources while the Class \texttt{5} languages are at the opposite end of the spectrum.\footnote{Because the language name to class list from \citeauthor{joshi-etal-2020-state} does not use an ISO 639-3 or Glottocode, we can only use the name to identify which language is paired with which Glottocode. We only assign classes for unambiguous language names. For example, while ``khmer'' is found in the language to class list, we do not join it with ``Central Khmer.'' There are 30 languages without an assigned Resource Class.}
The vast majority of missing outputs are concentrated in languages designated as Class 0 (e.g., Tamasheq, Nuer, Kabiyé), which represents the lowest-resource tier in our dataset. ``--'' means that the language was not found in the taxonomy.

Furthermore, the model fails far more frequently in the \emph{English-to-Target} direction (73 instances) than in the Target-to-English direction (7 instances). This indicates that the primary challenge is not the model's ability to process or analyze the target languages (i.e., $T \rightarrow E$), but rather its capacity to reliably \textit{generate} text in them (i.e., $E \rightarrow T$). This strongly suggests limited training data in the target language. This conclusion is reinforced by the performance on higher-resourced languages. We will now examine the overall translation quality of the outputs.

\paragraph{LLM performance is heavily influenced by translation direction, language family, and resource availability.} We use \textsc{chrF} \citep{popovic2015chrf} instead of \textsc{chrF+} or \textsc{chrF++} \citep{popovic2017chrf++} because the former is language independent and tokenization independent, which is needed when many languages found in FLORES-200 may not have a robust tokenizer or even have one readily available. \textsc{chrF} measures translation quality by calculating character-level n-gram overlap F-score between the machine translation and the human translation. The latter two introduces word unigram and bigram overlap into the equation.
We use the implementation provided by Hugging Face with default parameters,\footnote{\url{https://huggingface.co/spaces/evaluate-metric/chrf}} which adopts the implementation from sacreBLEU \citep{post-2018-call}\footnote{\url{https://github.com/mjpost/sacreBLEU\#chrf--chrf}} but with a slightly different input format.

\begin{table}[ht]
\centering
\begin{tabular}{lrr}
\toprule
\textbf{Direction} & \textbf{Mean \textsc{chrF} Score} & \textbf{Correlation with Class ($\rho$)} \\
\midrule
$E \rightarrow T$ & 43.92 & 0.598 \\
$T \rightarrow E$ & 64.27 & 0.466 \\
\bottomrule
\end{tabular}
\caption{Mean \textsc{chrF} scores and their Spearman's correlation ($\rho$) with resource class for each translation direction.}
\label{tab:summary_chrf_with_spearman}
\end{table}

\begin{figure}[htbp]

    \centering

    \includegraphics[width=0.95\textwidth]{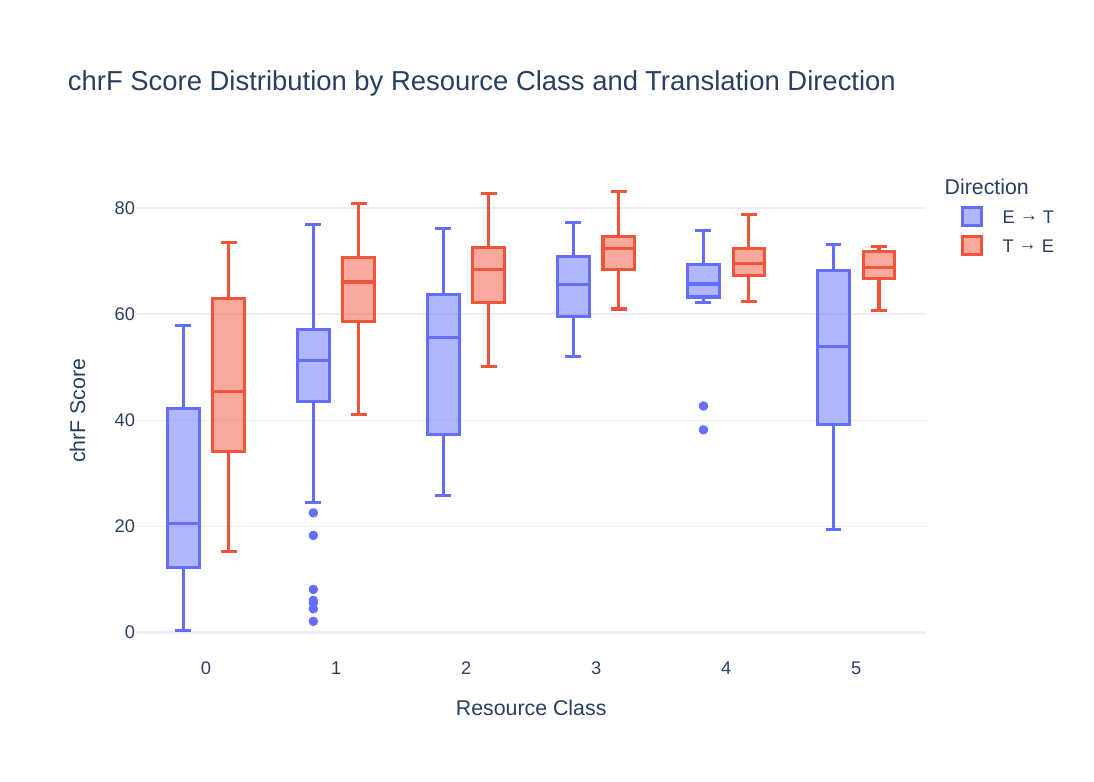}

    \caption{Comparison of \textsc{chrF} score distributions for English-to-Target ($E \rightarrow T$) and Target-to-English ($T \rightarrow E$) translations, grouped by resource class. The plot shows a clear positive trend where quality increases with resource availability, with the $T \rightarrow E$ direction consistently outperforming the $E \rightarrow T$ direction. Boxes represent the interquartile range, and points show individual languages that fall beyond the lower fence.}

    \label{fig:flores200_et_te_chrf_by_class}

\end{figure}

Worth noting is the direction where the model is worse on average ($E \rightarrow T$) is also the direction where performance is more strongly influenced by resource availability (higher correlation, $\rho = 0.598$). This suggests that while translating into English has a relatively high performance floor, the model's ability to generate text in other languages is both lower on average and more vulnerable to data scarcity. Figure ~\ref{fig:flores200_et_te_chrf_by_class} paints a similar picture in which lower resource classes predictably have worse performance compared to languages with more resources. We also see that translating from English to another language exacerbates the problem. 

\begin{table}[htbp]
\centering
\begin{tabular}{l rr rr}
\toprule
& \multicolumn{2}{c}{\textbf{$E \rightarrow T$}} & \multicolumn{2}{c}{\textbf{$T \rightarrow E$}} \\
\cmidrule(lr){2-3} \cmidrule(lr){4-5}
\textbf{Factor} & \textbf{Effect Size ($\eta_p^2$)} & \textbf{p-value} & \textbf{Effect Size ($\eta_p^2$)} & \textbf{p-value} \\
\midrule
Family & 0.409 & $< .001$ & 0.515 & $< .001$ \\
Class  & 0.412 & $< .001$ & 0.381 & $< .001$ \\
Script & 0.174 & .265     & 0.125 & .740     \\
\bottomrule
\end{tabular}
\caption{Summary of One-Way ANOVA results showing the influence of each factor on \textsc{chrF} scores. Effect sizes are given as partial eta-squared ($\eta_p^2$).}
\label{tab:anova_summary}
\end{table}

To also see how language family and script influence translation quality we used three separate one-way ANOVAs for each translation direction ($E \rightarrow T$ and $T \rightarrow E$). The results, summarized in Table~\ref{tab:anova_summary}, indicate that both \textbf{family} and \textbf{class} have a large and highly significant effect on performance in both directions (all $p < .001$). In contrast, \textbf{script} was not found to be a statistically significant predictor of \textsc{chrF} score in either analysis.

The analysis reveals an important asymmetry in the influence of resource class. While significant in both cases, \textbf{class} accounts for a larger portion of the variance in $E \rightarrow T$ scores ($\eta_p^2 = .412$) than in $T \rightarrow E$ scores ($\eta_p^2 = .381$).

This illustrates that processing low-resource languages still proves to be a challenge for even the most powerful of models. FLORES-200 only covers a small fraction of the world's languages and were chosen carefully based on several considerations, such as having a presence on Wikipedia. This limitation with processing low-resource languages will only be more pronounced when we examine other languages with even fewer resources. The results for each language can be found in Table~\ref{tab:flores200_language_level_results} as well as additional figures for script and language family-level scores in Section~\ref{appendix:supp-figures} of the Appendix.

Given that these results stem from a single experimental iteration, they should be interpreted as preliminary. Nevertheless, they provide strong evidence of the lopsided distribution of data resources among the world's languages and imbalanced performance across languages for today's SOTA LLMs, which warrants further investigation.

\section{Multi-Agent Framework for IOL Problems}
\label{section:multiagentframework}

\begin{figure}[htb!]
    \centering
        \includegraphics[width=1\textwidth]{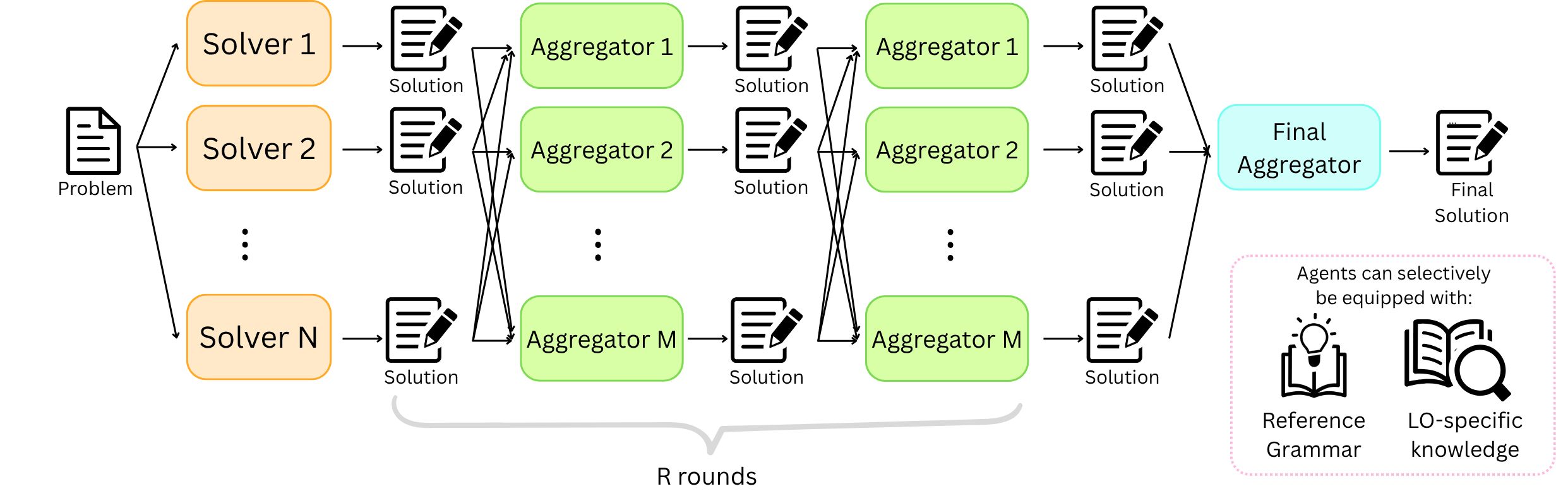}
        \caption{Multi-Agent Framework for Solving Linguistics Olympiad Problems}
        \label{fig:Framework}
\end{figure}

In this section, we introduce our proposed methodology for enhancing LLMs reasoning capabilities in linguistic problem solving.

\subsection{Multi-Agent Framework for IOL Problems}
\label{subsection:multiagentframework}

The framework consists of multiple interacting agents:

\begin{enumerate}
    \item \textbf{Solver Agent}: This agent proposes initial hypotheses regarding morphological, phonological, or syntactic structures based on provided linguistic data. 

    \item \textbf{Aggregator Agent}: Following Mixture-of-Agents (\cite{ICLR2025_5434be94}), this agent collects multiple proposed solutions, and was asked to generate its own solution.

    \item \textbf{Grammar Agent}: This agent utilizes a manually collected set of publicly available reference grammar books, each of which is annotated with the Glottocode. Given a problem, the agent would search inside the database to find the grammar book (if it exists), and summarize the relevant grammatical features or knowledge about the language. The idea that a grammar book may help the model deconstruct a language aligns with \cite{ICLR2024_52d63f9e}. The implementation is detailed in Section~\ref{subsubsec:grammar_agent}.
\end{enumerate}

\subsubsection{Grammar Agent}\label{subsubsec:grammar_agent}
Our Grammar Agent makes use of a knowledge base to enable retrieval-augmented generation (RAG) \citep{lewis2020rag} of linguistic reference books, with the majority being reference grammars. These materials were gathered from publicly available resources online. Each reference book covers a specific language and was manually annotated with its corresponding Glottocode. This allows us to include rich linguistic metadata to assist our agent in searching or solving problems.

While some files already have text embedded in the document, others need to be processed with OCR first. We use Mistral's OCR API to complete this task.\footnote{\url{https://mistral.ai/news/mistral-ocr}}. The output is a list containing Markdown for each page in a file. For each file, we concatenate all Markdown pages into one large Markdown file. 

To enable vector search, we first chunk our texts into lengths of 256 tokens. We use \textsc{Qwen/Qwen3-Embedding-4B} to tokenize and to subsequently embed our chunks.\footnote{We tokenize via Hugging Face's Transformers library while we use DeepInfra's API to generate embeddings: \url{https://deepinfra.com/Qwen/Qwen3-Embedding-4B/api}.}

For the files with text already embedded, we use Docling \citep{Docling} to parse the structure of each file and prepend contextual information within each chunk, via the `contextualize()` method available with the \texttt{HybridChunker}. 
For the Markdown files, we use Unstructured\footnote{\url{https://github.com/Unstructured-IO/unstructured/releases/tag/0.18.3}} to automatically parse the structure and to prepend the titles of sections to each chunk.

After all texts have been chunked, we use the aforementioned embedding API to vectorize all of our chunks. We use LanceDB\footnote{\url{https://github.com/lancedb/lancedb/releases/tag/python-v0.24.1}} as our vector database because of its simpler design as a serverless database as well as support for hybrid search and using SQL queries directly for performing more advanced searches. 
Our Grammar Agent can not only search for relevant texts via full-text search, vector search, and hybrid search, it can also within metadata fields, such specific Glottocodes, language families, languoid names, and macroareas and countries that speak the language. Our final knowledge base includes data covering over 1100 languages across over 140 language families. 

\subsection{Results and Analysis}
\label{sec:agent-result-and-analysis}

We conducted a set of preliminary experiments to evaluate our multi-agent framework against several baselines. The results are summarized in Table~\ref{tab:agent_performance_summary} with experiments including:

\begin{itemize}

\item Vanilla baseline: A direct, single-pass call to an LLM (OpenAI-o4-mini and Gemini-2.5-pro) to solve the problem, following the required output format. We used OpenAI-o4-mini and Gemini-2.5-pro for the experiments, with temperature set to 0.75.
\item Guided prompt: A major drawback of the vanilla prompting is that, usually the LLM is not familiar with the underlying assumptions about Linguistics Puzzles (e.g., ``All the questions are self-contained'', ``The final solution should be able to explain 100\% of the examples, not just 90\%''). To inform the model about such nuances, we include the contents of the book \textit{Linguistics Olympiad: Training guide} \citep{Neacșu2024} into the system prompt.
\item Grammar agent: A Grammar Agent is employed (as mentioned in Section \ref{subsection:multiagentframework}). Since the reference grammar database does not cover all languages, only problems with a reference are counted in this setting. It should be noted that the relationship between the language coverage of reference grammar books and that of the problems remains unexplored. To ensure a fairer comparison, the baseline scores should include the same set of problems. See Appendix~\ref{tab:agent_performance_summary_subset})
\item Single agent, multi-rounds: The solution of a solver will be fed into itself for multiple rounds. Equivalent to the Mixture-of-Agent setting with N=M=1.
\item Mixture-of-Agents: A multi-round framework as depicted in Figure~\ref{fig:Framework}. We used 2 agents for each layer (following the notation in the figure, N=2) and tested with from 0 to 4 extra layers of fully connected aggregators (M=2, R $\in \{0,1,2,3, 4\}$), in addition to the final aggregator.

\end{itemize}

\begin{table}[htbp]
\centering
\resizebox{\textwidth}{!}{%
\begin{tabular}{lrrrrrr}
\toprule
\textbf{Run ID} & \textbf{Avg Score} & \textbf{Scores $\in \boldsymbol{[0, 0.25)}$} & \textbf{$\boldsymbol{[0.25, 0.5)}$} & \textbf{$\boldsymbol{[0.5, 0.75)}$} & \textbf{$\boldsymbol{[0.75, 1]}$} & \textbf{Total} \\
\midrule
OpenAI-o4-mini (baseline) & 0.193 & 61 & 18 & 4 & 6 & 89 \\
Gemini-2.5-pro (baseline) & 0.381 & 38 & 23 & 20 & 14 & 95 \\
\midrule
OpenAI-o4-mini (guided) & 0.194 & 63 & 16 & 5 & 5 & 89 \\
Gemini-2.5-pro (guided) & 0.307 & 45 & 22 & 13 & 11 & 91 \\
\midrule
Gemini-2.5-pro (w/ grammar agent) & 0.387 & 30 & 18 & 17 & 12 & 77 \\
\midrule
Gemini-2.5-pro (Single agent, $1^{st}$ round)\textsuperscript{\textdagger} & 0.353 & 39 & 24 & 20 & 12 & 95 \\
Gemini-2.5-pro (Single agent, 2 rounds) & 0.373 & 40 & 21 & 19 & 15 & 95 \\
Gemini-2.5-pro (Single agent, 3 rounds) & 0.380 & 38 & 22 & 21 & 14 & 95 \\
Gemini-2.5-pro (Single agent, 4 rounds) & 0.383 & 40 & 19 & 21 & 15 & 95 \\
Gemini-2.5-pro (Single agent, 5 rounds) & 0.384 & 40 & 20 & 20 & 15 & 95 \\
\midrule
OpenAI-o4-mini (Single agent, $1^{st}$ round)\textsuperscript{\textdagger} & 0.186 & 66 & 14 & 4 & 6 & 90 \\
OpenAI-o4-mini (Single agent, 2 rounds) & 0.185 & 69 & 13 & 4 & 6 & 92 \\
OpenAI-o4-mini (Single agent, 3 rounds) & 0.188 & 67 & 15 & 5 & 6 & 93 \\
OpenAI-o4-mini (Single agent, 4 rounds) & 0.199 & 64 & 15 & 6 & 6 & 91 \\
OpenAI-o4-mini (Single agent, 5 rounds) & 0.183 & 67 & 14 & 4 & 6 & 91 \\
\midrule
Gemini-2.5-pro (MoA, $1^{st}$ round)\textsuperscript{\textdagger} & 0.410 & 39 & 15 & 24 & 17 & 95 \\
Gemini-2.5-pro (MoA, R=0, (2 rounds)) & 0.425 & 36 & 19 & 21 & 19 & 95 \\
Gemini-2.5-pro (MoA, R=1, (3 rounds)) & 0.453 & 30 & 21 & 23 & 21 & 95 \\
Gemini-2.5-pro (MoA, R=2, (4 rounds)) & 0.449 & 29 & 21 & 25 & 20 & 95 \\
Gemini-2.5-pro (MoA, R=3, (5 rounds)) & 0.458 & 28 & 22 & 23 & 22 & 95 \\
Gemini-2.5-pro (MoA, R=4, (6 rounds)) & 0.459 & 28 & 23 & 21 & 22 & 94 \\
\midrule
OpenAI-o4-mini (MoA, first round)\textsuperscript{\textdagger} & 0.172 & 65 & 15 & 7 & 4 & 91 \\
OpenAI-o4-mini (MoA, R=0 (2 rounds)) & 0.319 & 46 & 22 & 11 & 12 & 91 \\
OpenAI-o4-mini (MoA, R=1 (3 rounds)) & 0.383 & 38 & 20 & 18 & 15 & 91 \\
OpenAI-o4-mini (MoA, R=2 (4 rounds)) & 0.384 & 38 & 22 & 21 & 14 & 95 \\
OpenAI-o4-mini (MoA, R=3 (5 rounds)) & 0.397 & 36 & 23 & 20 & 14 & 93 \\
OpenAI-o4-mini (MoA, R=4 (6 rounds)) & 0.417 & 36 & 17 & 24 & 16 & 93 \\

\bottomrule
\end{tabular}%
}
\caption{Summary of agent performance, showing average scores and the distribution of problem scores. Each row represents a unique experimental setting. For the results with multiple rounds, the name denotes the model used in the final layer (i.e, the final solution is generated by it). The rows marked with a dagger (\textdagger) means that its setting is equivalent to the baseline, and therefore the score differences demonstrate model stochasticity.}
\label{tab:agent_performance_summary}
\end{table}
Note that the (preliminary) results are to be interpreted with caution, as each experiment was conducted only once. Given the inherent stochasticity of LLMs, statistical tests on multiple runs would be required to make definitive claims about the efficacy of different settings.

Nonetheless, we observe a general trend consistent with \citet{ICLR2025_5434be94}, namely, that increasing the number of rounds helped. For both Gemini-2.5-pro and OpenAI-o4-mini, the average score in the Mixture-of-Agents (MoA) setting consistently increases with more rounds. A similar, though less pronounced, trend is visible in the single-agent multi-round setting for Gemini-2.5-pro. In contrast, the effectiveness of the guided prompt and the Grammar Agent is less clear. For example, the guided setting for Gemini-2.5-pro scored lower than its baseline. However, as noted in Table~\ref{tab:agent_performance_summary}, the baseline and single-agent (1-round) settings, which are conceptually identical, also show significant variance. This run-to-run instability makes it difficult to attribute performance changes solely to these specific interventions without more controlled experiments.

The total number of problems graded for each setting was slightly fewer than that of the benchmark (96 in total), because the model might not follow the format consistently. The format-following abilities appear positively correlated with the average score of the system (with the exception of the Grammar Agent setting, where problems were selectively included).

These results underscore the need for more rigorous studies. The apparent benefits of the MoA framework deserve further investigation via ablation studies to disentangle the effects of parallel generation (multi) from iterative refinement (round). Future work should aim to isolate the contribution of the Aggregator Agent: does it primarily select the best solution from the previous round, or does it perform novel reasoning by synthesizing multiple inputs? Because of the difference between the settings (e.g., Large Reasoning Models vs. LLMs), there is no strong evidence to suggest that the trend will follow past MoA discussions such as \cite{li2025rethinkingmixtureofagentsmixingdifferent}. Answering such questions is key to understanding and optimizing such agentic architectures.




\section{Conclusion}

In this work, we introduced LingBench++, a linguistically-informed benchmark designed to move beyond final-answer accuracy and enable a granular assessment of an LLM's reasoning on complex linguistic structures. Our typological analysis of IOL problems provides a structured lens for this evaluation, while our empirical study of a state-of-the-art model on the FLORES-200 dataset underscored the critical need for improved cross-linguistic generalization, particularly in low-resource settings. The benchmark will be publicly released, and we call on the community to build on this foundation to look inward at the nascent logic of LLMs, and outward at the boundless diversity of language that inspires them.

\section*{Acknowledgement}
We thank Hung-Chi Chen, Yin-Shuo Chang, Kanoa Ziyang Teng and Chloe Cheng and others for the annotation and refinement of the data.

\bibliographystyle{apalike}
\bibliography{iol-ai.bib}




\newpage
\appendix

\section{UKLO Classification Framework for Problems}
\label{appendix:uklo-classification}
\begin{itemize}
    \item \textbf{Subjects} – For a given subject to appear in the classification, at least two rules in the solution must be of that type.
    \begin{itemize}
        \item \textbf{Compounding}: The problems mainly focus on deducing the dictionary meanings of words by analyzing how the meaning changes when different word components are combined.
        
        \item \textbf{Morphology}: The problems primarily require understanding how morphemes (the smallest units of meaning) combine to form grammatical words.
        
        \item \textbf{Numbers}: The problems are centered on understanding the structure and formation of numerals and numeral expressions. 
        
        \item \textbf{Phonology and Phonetics}: The problems focus on the sounds of a language and how they are organized. Phonology deals with sound systems within specific languages and in general, while phonetics studies the nature, production, and perception of speech sounds, independent of any particular language.
        
        \item \textbf{Semantics}: The problems emphasize understanding how meaning influences language, especially how meaning shapes grammar and how different languages express the same concepts with different words.
        
        \item \textbf{Syntax}: The problems focus on understanding how words combine to form phrases and sentences.
        
        \item \textbf{Writing System}: The problems involve analyzing writing systems, including both the use of the Latin alphabet in various languages and other scripts.
    \end{itemize}
    
    \item \textbf{Problem Type}
    \begin{itemize}
        \item \textbf{Rosetta}: The problems consist of sets of corresponding words or phrases across different languages or writing systems, with most pairings provided. Some elements may be missing, creating gaps that need to be filled. Solving the task requires generating new correspondences, typically translations.
        
        \item \textbf{Match-up}: The problems consist of sets of corresponding words or phrases across multiple languages or writing systems, with only a few pairings given. Some words may not belong to any set, but it still qualifies as a match-up. The task involves identifying new correspondences, usually translations.
        
        \item \textbf{Monolingual}: The problems are texts in an unfamiliar language (or equivalent writing system), generally without direct translations or transliterations, except perhaps for one or two words. To solve the task, you must translate the text from the unknown language.
        
        \item \textbf{Pattern}: The problems consist of words or groups of word forms or cognates that follow a certain pattern, though there may be exceptions. To solve the task, you must either generate other examples that fit the pattern or identify exceptions, without relying on translation as in Rosetta tasks.
        
        \item \textbf{Computational}: The problems include a description of a computational or logical system. Solving the problem involves analyzing and implementing the system according to the given rules.
        
        \item \textbf{Text}: The problems consist of full texts in different languages or scripts, without being broken down into smaller parts. To solve the task, you must infer linguistic rules using context and other cues.
    \end{itemize}
    
    \item \textbf{Theme}
    \begin{itemize}
        \item \textbf{Classical}: These problems feature languages that were primarily spoken around a thousand years ago or earlier.
        
        \item \textbf{Comparative}:These problems involve comparing either related languages or different historical stages of a single language.
        
        \item \textbf{Encrypted}: These problems involve deciphering an encoded message in English.
        
        \item \textbf{Kinship}: These problems focus on understanding how different languages and cultures describe family relationships and naming systems.
        
        \item \textbf{Maps}: These problems explore how various languages express and conceptualize directions and spatial orientation.
        
        \item \textbf{Mystery}: These problems include a mystery element that draws on general or world knowledge, often involving content beyond linguistics.
        
        \item \textbf{MFL}: These problems involve languages commonly taught in secondary school modern foreign language (MFL) departments, or closely related languages (e.g., those from the Romance or Germanic families).
        
        \item \textbf{Senses and Feelings}: These problems examine linguistic expressions related to emotions or sensory experiences (e.g., smells, sounds).
        
        \item \textbf{Stories}: These problems either contain a narrative storyline or feature one or more fictional characters. They use storytelling to create engaging contexts for linguistic analysis, often drawing from literary traditions.
        
        \item \textbf{Poetry}: These problems revolve around the structure and features of poetic language.
        
        \item \textbf{No Theme (N/A)}: These problems focus on core linguistic topics without any specific thematic context.
    \end{itemize}
\end{itemize}

\newpage

\section{Example of Kinship Problems}
\label{appendix:kinship-example}
\begin{figure*}[htbp!]
    \centering
    \begin{minipage}[ht]{0.8\textwidth}
        \centering
        \includegraphics[width=\linewidth]{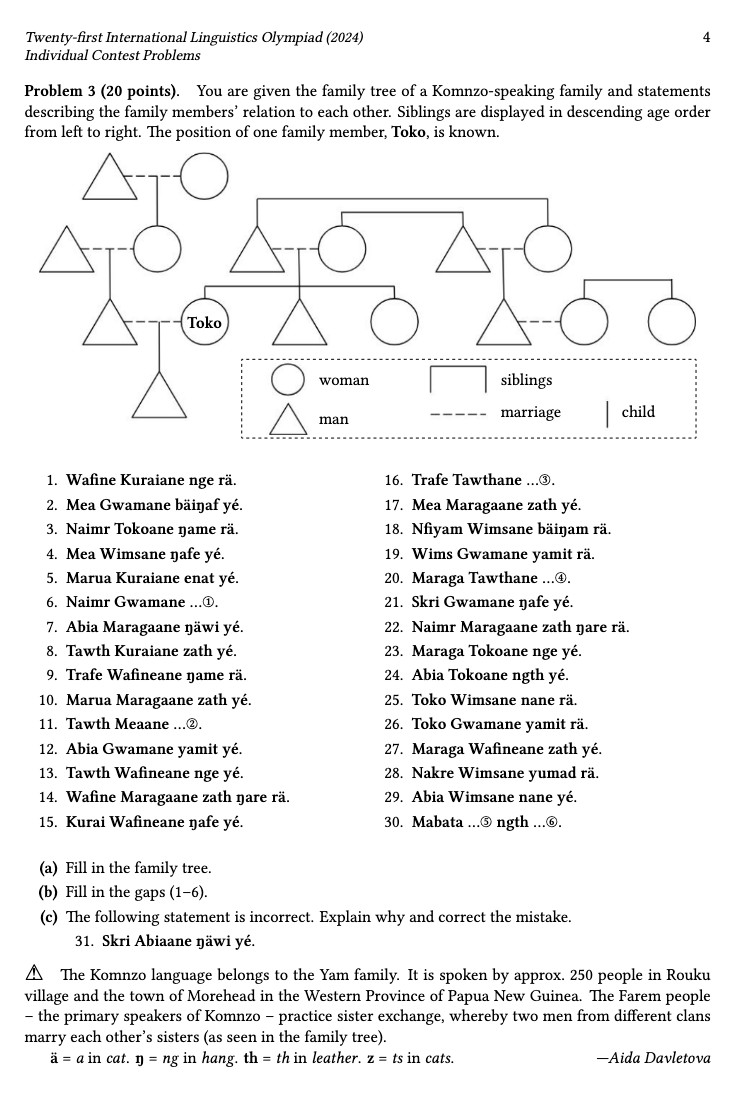}
        \caption{Original Problem 3 in 2024.}
        \label{fig:kinship}
    \end{minipage}
\end{figure*}
    \hfill
    \vspace{0.5cm}
    \begin{minipage}[t]{\textwidth}
        \textbf{Transcription of the Family Tree}
        \begin{itemize}
    \item Man 1 and Woman 1 are married. Their child is Woman 2.
    \item Man 2 and Woman 2 are married. Their child is Man 3.
    \item Man 3 and Woman 3 are married. Their child is Man 4.
    \item Woman 3 is Toko.
    \item Man 5 and Woman 4 are married. Their children are Woman 3, Man 6 and Woman 5, from oldest to youngest.
    \item Man 5 and Woman 6 are siblings. The former is older.
    \item Woman 4 and Man 7 are siblings. The former is older.
    \item Man 7 and Woman 6 are married. Their child is Man 8.
    \item Man 8 and Woman 7 are married.
    \item Woman 7 and Woman 8 are siblings. The former is older.
\end{itemize}

    \end{minipage}

\newpage

\section{Prompt Template for Reasoning Process Generation}
\label{appendix:prompt-template}

The following Python template was used to generate reasoning chains for IOL problems:

\begin{lstlisting}[language=Python]
## Prompt:
As an expert in linguistics solve the following problem. Given the following IOL problem and its answer, generate a detailed, step-by-step chain of thoughts that could specifically and reasonably lead to the answer. Focus on the reasoning process, essential linguistic rules, logical deductions, and the final solution. Make your whole output into a markdown file.

## Problem:
{problem_text}

## Solution:
{solution_text}

## Your response:

\end{lstlisting}

\newpage

\section{System Prompt for Model Reasoning Evaluation}
\label{appendix:eval-prompt-metric-example}
\begin{lstlisting}[language=Python, label={lst:system_prompt}]
system_prompt = """Given the evaluation rules and metrics for model reasoning of IOL problems, consider the golden reasoning reference, and evaluate the target model reasoning with the metrics of five dimensions. 
evaluation rules and metrics (5-score):
{metrics}

scoring_:
{scoring}

golden reasoning reference:
{golden_reasoning_reference}

target model reasoning:
{model_reasoning}
"""

metrics = """
### Metrics and Descriptions (Bullet Points)
 (i) 3.1 Information Extraction & Structuring
  * **Stepwise Logical Validity Score (SLVS)**: Measures whether each reasoning step is logically valid and aligned with the golden reasoning reference (GRR).
  * **Information Structuring Completeness (ISC)**: Measures completeness of extracted key information and its structure compared to GRR.
  ... [TRUNCATED FOR BREVITY IN PAPER]
"""

scoring = """
## Reasoning Quality Evaluation -- Scoring Rubric (5 Points per Metric)

| **Dimension** | **Metric** | **Score 5 (Excellent)** | **Score 3 (Acceptable)** | **Score 1 (Poor)** |
| ----- | ----- | ----- | ----- | ----- |
| **(i) 3.1 Information Extraction & Structuring** | **SLVS**   | All reasoning steps are logically valid and follow GRR structure | Minor logical flaws or omissions; generally coherent | Major logical errors, incoherent or illogical steps | |  | **ISC**    | Extracts and structures all key information as per GRR | Extracts partial or incomplete key information | Fails to extract/structure key information |
... [TRUNCATED FOR BREVITY IN PAPER]
"""

golden_reasoning_reference = """
# Your response:

# Chain of Thought: Solving the Swift News Linguistics Problem

... [TRUNCATED FOR BREVITY IN PAPER]
"""

target_model_reasoning = """
**Solving the Linguistic Puzzle**

... [TRUNCATED FOR BREVITY IN PAPER]
"""
\end{lstlisting}

\newpage

\section{Resolution of Ambiguous ISO 639-3 to Glottocode Mappings}

\begin{table}[ht]
\centering
\caption{Resolution of ambiguous source ISO 639-3 codes to specific language varieties and their corresponding Glottocode.}
\label{tab:iso-glottolog-mapping}
\begin{tabular}{l p{0.9\textwidth}}
\toprule
\multicolumn{2}{l}{\textbf{Language Mapping Details}} \\
\midrule

\multirow{3}{*}{\texttt{srd}} & \textbf{Language:} Sardinian \\
 & \textbf{ISO $\rightarrow$ Glottocode:} None $\rightarrow$ \texttt{sard1257} \\
 & \textbf{Justification:} Top-level family node. \\
\midrule

\multirow{3}{*}{\texttt{est}} & \textbf{Language:} Estonian \\
 & \textbf{ISO $\rightarrow$ Glottocode:} \texttt{ekk} $\rightarrow$ \texttt{esto1258} \\
 & \textbf{Justification:} Primary language entry. \\
\midrule

\multirow{3}{*}{\texttt{kon}} & \textbf{Language:} South-Central Kongo \\
 & \textbf{ISO $\rightarrow$ Glottocode:} \texttt{kng} $\rightarrow$ \texttt{koon1244} \\
 & \textbf{Justification:} Known as Kongo in World Atlas of Language Structures (WALS). \\
\midrule

\multirow{3}{*}{\texttt{zho}} & \textbf{Language:} Mandarin \\
 & \textbf{ISO $\rightarrow$ Glottocode:} \texttt{cmn} $\rightarrow$ \texttt{mand1415} \\
 & \textbf{Justification:} Most populous variety. \\
\midrule

\multirow{3}{*}{\texttt{grn}} & \textbf{Language:} Eastern Bolivian Guaraní \\
 & \textbf{ISO $\rightarrow$ Glottocode:} \texttt{gui} $\rightarrow$ \texttt{east2555} \\
 & \textbf{Justification:} Guaraní categorized as Class 1 in \citet{joshi-etal-2020-state}, which aligns more with Ethnologue's Digital Language Support classification of ``Ascending'' for the language. \\
\bottomrule
\end{tabular}
\end{table}

\newpage

\section{Preliminary Analysis of IOL Problems.}
\label{appendix:preliminary-analysis}
\begin{figure}[htbp!]
    \centering
    \begin{subfigure}[b]{0.48\textwidth}
        \includegraphics[width=\textwidth]{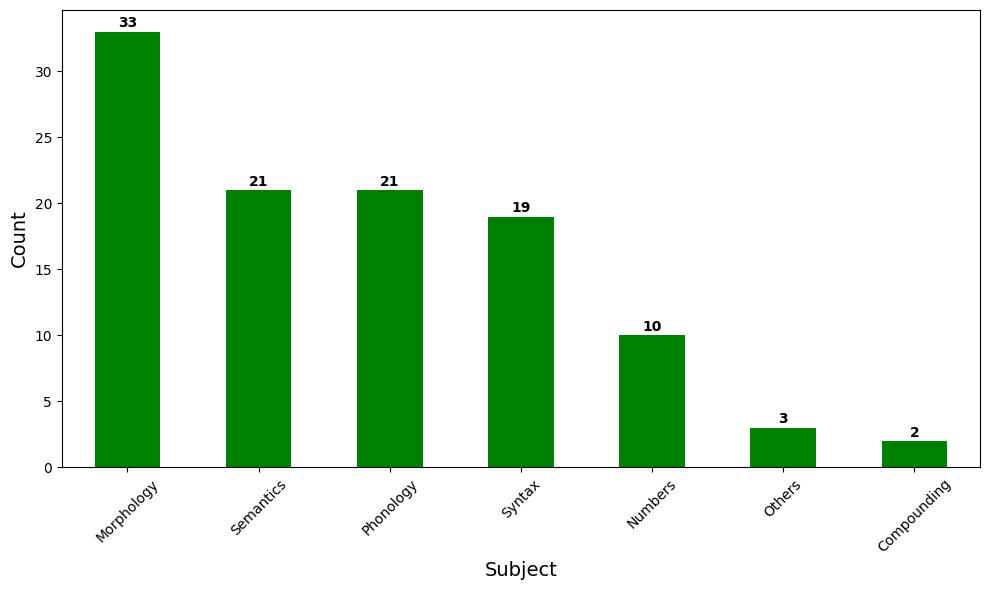}
        \caption{Subject Distribution}
        \label{fig:sub1}
    \end{subfigure}
    \hfill
    \begin{subfigure}[b]{0.48\textwidth}
        \includegraphics[width=\textwidth]{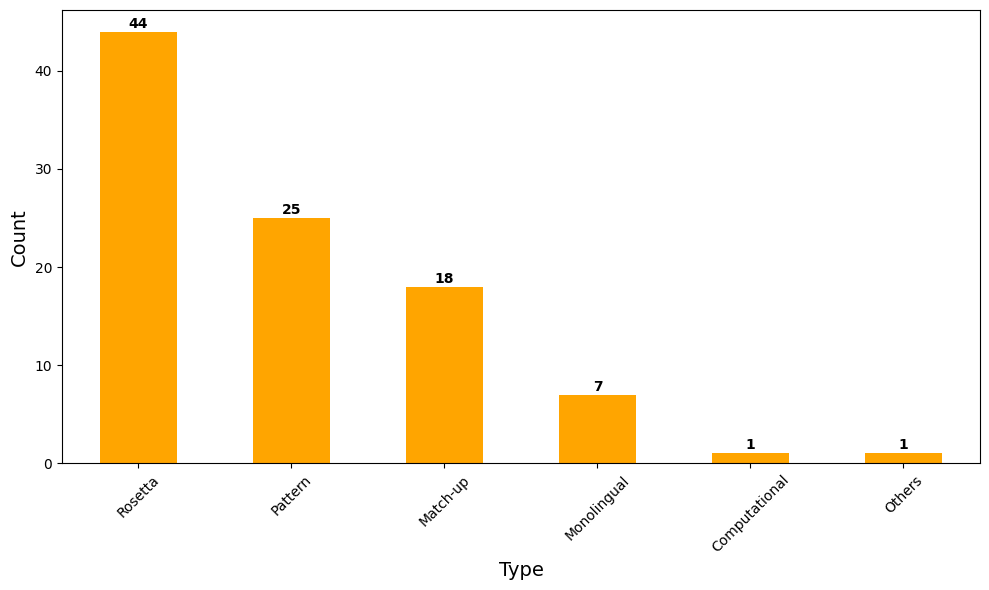}
        \caption{Type Distribution}
        \label{fig:sub2}
    \end{subfigure}
    
    \vspace{0.5cm} 
    
    \begin{subfigure}[b]{0.48\textwidth}
        \includegraphics[width=\textwidth]{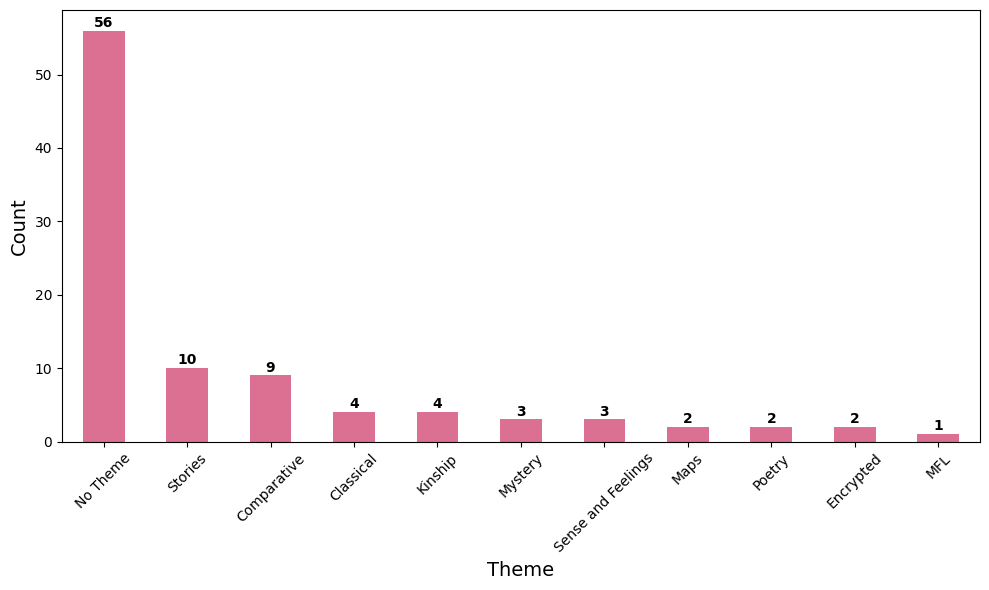}
        \caption{Theme Distribution}
        \label{fig:sub3}
    \end{subfigure}
    \hfill
    \begin{subfigure}[b]{0.48\textwidth}
        \includegraphics[width=\textwidth]{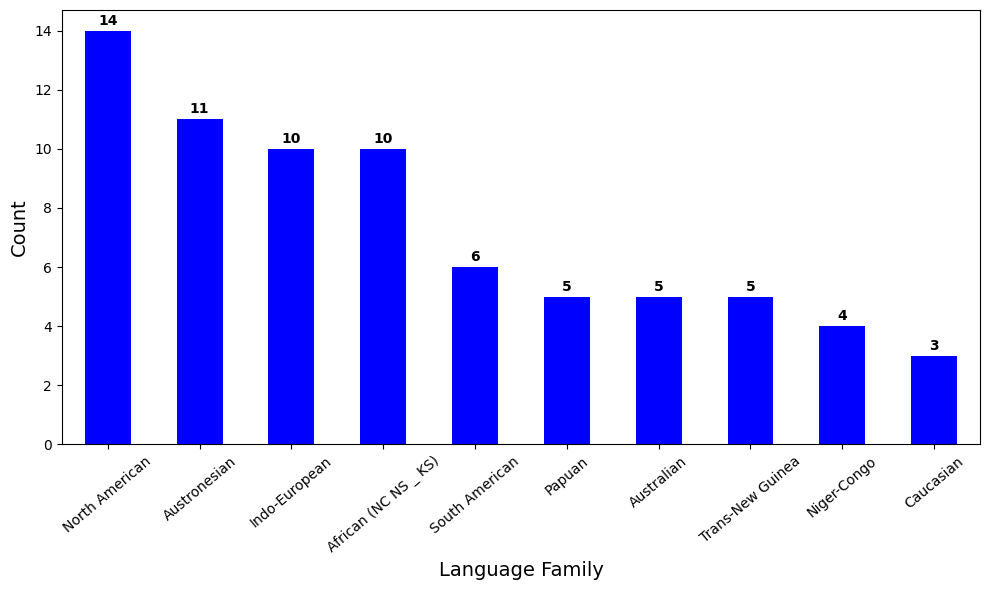}
        \caption{Language Family Distribution}
        \label{fig:sub4}
    \end{subfigure}
     \caption{Statistical distributions of various features in the IOL problems dataset.}
    \label{fig:iol-problem-stats}
\end{figure}

\newpage
\section{Language-Level \textsc{chrF} Translation Scores for Gemini-2.5-Flash on FLORES-200}

{
\footnotesize 
\begin{longtable}{p{6cm}cclcc}

\caption{Performance results by language, including \textsc{chrF} scores, sample counts, and resource class.}
\label{tab:flores200_language_level_results} \\

\toprule
\multicolumn{1}{m{4cm}}{\centering\textbf{Language \\ (glottocode\_Script)}} &
\multicolumn{1}{m{1.5cm}}{\centering\textbf{$E \rightarrow T$ \\ \textsc{chrF}}} &
\multicolumn{1}{m{1.5cm}}{\centering\textbf{$T \rightarrow E$ \\ \textsc{chrF}}} &
\textbf{Family} &
\textbf{Class} &
\multicolumn{1}{m{2.5cm}}{\centering\textbf{Samples \\ ($E \rightarrow T$ / $T \rightarrow E$)}} \\
\midrule
\endfirsthead

\multicolumn{6}{c}%
{{\tablename\ \thetable{} -- continued from previous page}} \\
\toprule
\multicolumn{1}{m{4cm}}{\centering\textbf{Language \\ (glottocode\_Script)}} &
\multicolumn{1}{m{1.5cm}}{\centering\textbf{$E \rightarrow T$ \\ \textsc{chrF}}} &
\multicolumn{1}{m{1.5cm}}{\centering\textbf{$T \rightarrow E$ \\ \textsc{chrF}}} &
\textbf{Family} &
\textbf{Class} &
\multicolumn{1}{m{2.5cm}}{\centering\textbf{Samples \\ ($E \rightarrow T$ / $T \rightarrow E$)}} \\
\midrule
\endhead

\midrule
\multicolumn{6}{r}{{Continued on next page}} \\
\endfoot

\bottomrule
\endlastfoot
Acehnese (achi1257\_Arabic) & 6.05 & 49.46 & Austronesian & 1 & 9 / 10 \\
Acehnese (achi1257\_Latin) & 46.42 & 71.11 & Austronesian & 1 & 10 / 10 \\
Afrikaans (afri1274\_Latin) & 73.91 & 83.15 & Indo-European & 3 & 10 / 10 \\
Akan (akan1250\_Latin) & 37.78 & 49.00 & Atlantic-Congo & 1 & 10 / 10 \\
Amharic (amha1245\_Ethiopic (Ge`ez)) & 35.85 & 70.71 & Afro-Asiatic & 2 & 10 / 10 \\
Assamese (assa1263\_Bengali) & 48.08 & 67.72 & Indo-European & 1 & 10 / 10 \\
Asturian-Leonese-Cantabrian (astu1245\_Latin) & 69.93 & 73.94 & Indo-European & 1 & 10 / 10 \\
Awadhi (awad1243\_Devanagari (Nagari)) & 41.45 & 67.04 & Indo-European & 0 & 10 / 10 \\
Ayacucho Quechua (ayac1239\_Latin) & 37.25 & 53.34 & Quechuan & -- & 9 / 10 \\
Balinese (bali1278\_Latin) & 44.79 & 61.53 & Austronesian & 0 & 10 / 10 \\
Bambara (bamb1269\_Latin) & 2.12 & 41.72 & Mande & 1 & 8 / 10 \\
Banjar (banj1239\_Arabic) & 4.46 & 53.69 & Austronesian & 1 & 10 / 10 \\
Banjar (banj1239\_Latin) & 51.99 & 60.64 & Austronesian & 1 & 10 / 10 \\
Bashkir (bash1264\_Cyrillic) & 56.01 & 68.67 & Turkic & 1 & 10 / 10 \\
Basque (basq1248\_Latin) & 64.81 & 67.00 & Unknown & 4 & 10 / 10 \\
Belarusian (bela1254\_Cyrillic) & 52.41 & 60.98 & Indo-European & 3 & 10 / 10 \\
Bemba (Zambia) (bemb1257\_Latin) & 43.05 & 60.86 & Atlantic-Congo & 0 & 10 / 10 \\
Bengali (beng1280\_Bengali) & 59.45 & 68.50 & Indo-European & 3 & 10 / 10 \\
Bhojpuri (bhoj1244\_Devanagari (Nagari)) & 44.14 & 62.46 & Indo-European & 1 & 10 / 10 \\
Bosnian Standard (bosn1245\_Latin) & 67.72 & 70.89 & Indo-European & 3 & 10 / 10 \\
Buginese (bugi1244\_Latin) & 35.98 & 48.74 & Austronesian & 1 & 10 / 10 \\
Bulgarian (bulg1262\_Cyrillic) & 76.45 & 76.70 & Indo-European & 3 & 10 / 10 \\
Burmese (nucl1310\_Myanmar (Burmese)) & 53.93 & 68.35 & Sino-Tibetan & 1 & 10 / 10 \\
Catalan (stan1289\_Latin) & 67.90 & 72.33 & Indo-European & 4 & 10 / 10 \\
Cebuano (cebu1242\_Latin) & 65.84 & 80.13 & Austronesian & 3 & 10 / 10 \\
Central Aymara (cent2142\_Latin) & 31.09 & 44.91 & Aymaran & -- & 9 / 10 \\
Central Kanuri (cent2050\_Arabic) & 2.31 & 15.26 & Saharan & 0 & 10 / 8 \\
Central Kanuri (cent2050\_Latin) & 8.76 & 32.46 & Saharan & 0 & 6 / 10 \\
Central Khmer (cent1989\_Khmer) & 43.45 & 73.44 & Austroasiatic & -- & 10 / 10 \\
Central Kurdish (cent1972\_Arabic) & 51.29 & 67.85 & Indo-European & -- & 10 / 10 \\
Central Moroccan Berber (cent2194\_Tifinagh (Berber)) & 26.34 & 45.68 & Afro-Asiatic & 0 & 10 / 10 \\
Chhattisgarhi (chha1249\_Devanagari (Nagari)) & 50.58 & 70.19 & Indo-European & -- & 10 / 10 \\
Chokwe (chok1245\_Latin) & 19.24 & 28.74 & Atlantic-Congo & -- & 8 / 9 \\
Crimean Tatar (crim1257\_Latin) & 45.90 & 70.46 & Turkic & 1 & 10 / 10 \\
Croatian Standard (croa1245\_Latin) & 62.14 & 69.88 & Indo-European & 4 & 10 / 10 \\
Czech (czec1258\_Latin) & 63.42 & 73.96 & Indo-European & 4 & 10 / 10 \\
Danish (dani1285\_Latin) & 77.23 & 75.40 & Indo-European & 3 & 10 / 10 \\
Dari (dari1249\_Arabic) & 42.70 & 65.11 & Indo-European & 4 & 10 / 10 \\
Dutch (dutc1256\_Latin) & 66.63 & 68.29 & Indo-European & 4 & 10 / 10 \\
Dyula (dyul1238\_Latin) & 16.31 & 33.30 & Mande & 0 & 8 / 10 \\
Dzongkha (dzon1239\_Tibetan) & 33.37 & 50.97 & Sino-Tibetan & 1 & 9 / 10 \\
East Latvian (east2282\_Latin) & 45.29 & 73.02 & Indo-European & -- & 10 / 10 \\
Eastern Armenian (nucl1235\_Armenian) & 61.44 & 71.83 & Indo-European & 1 & 10 / 10 \\
Eastern Panjabi (panj1256\_Gurmukhi) & 56.52 & 73.75 & Indo-European & -- & 10 / 10 \\
Eastern Yiddish (east2295\_Hebrew) & 42.70 & 83.12 & Indo-European & -- & 10 / 10 \\
Egyptian Arabic (egyp1253\_Arabic) & 52.11 & 65.85 & Afro-Asiatic & 3 & 10 / 10 \\
Esperanto (espe1235\_Latin) & 66.90 & 76.29 & Artificial Language & 1 & 10 / 10 \\
Estonian (esto1258\_Latin) & 61.18 & 67.24 & Uralic & 3 & 10 / 10 \\
Ewe (ewee1241\_Latin) & 36.73 & 49.73 & Atlantic-Congo & 1 & 9 / 10 \\
Faroese (faro1244\_Latin) & 64.14 & 77.47 & Indo-European & 1 & 10 / 10 \\
Fijian (fiji1243\_Latin) & 50.32 & 55.60 & Austronesian & 1 & 10 / 10 \\
Finnish (finn1318\_Latin) & 66.57 & 68.23 & Uralic & 4 & 10 / 10 \\
Fon (fonn1241\_Latin) & 7.64 & 23.20 & Atlantic-Congo & 0 & 5 / 10 \\
French (stan1290\_Latin) & 73.10 & 70.06 & Indo-European & 5 & 10 / 10 \\
Friulian (friu1240\_Latin) & 61.78 & 67.92 & Indo-European & 1 & 10 / 10 \\
Galician (gali1258\_Latin) & 65.03 & 70.14 & Indo-European & 3 & 10 / 10 \\
Ganda (gand1255\_Latin) & 42.86 & 56.15 & Atlantic-Congo & 1 & 10 / 10 \\
Georgian (nucl1302\_Georgian (Mkhedruli)) & 56.55 & 63.11 & Kartvelian & 3 & 10 / 10 \\
German (stan1295\_Latin) & 71.48 & 72.08 & Indo-European & 5 & 10 / 10 \\
Gilit Mesopotamian Arabic (meso1252\_Arabic) & 51.31 & 66.27 & Afro-Asiatic & -- & 10 / 10 \\
Guarani (east2555\_Latin) & 30.45 & 60.51 & Tupian & 1 & 9 / 10 \\
Gujarati (guja1252\_Gujarati) & 49.30 & 70.17 & Indo-European & 1 & 10 / 10 \\
Haitian (hait1244\_Latin) & 62.81 & 69.51 & Indo-European & 2 & 10 / 10 \\
Halh Mongolian (halh1238\_Cyrillic) & 54.87 & 71.44 & Mongolic-Khitan & 0 & 10 / 10 \\
Hausa (haus1257\_Latin) & 61.93 & 67.27 & Afro-Asiatic & 2 & 10 / 10 \\
Hausa States Fulfulde (nige1253\_Latin) & 23.36 & 34.24 & Atlantic-Congo & -- & 10 / 10 \\
Hindi (hind1269\_Devanagari (Nagari)) & 64.11 & 69.33 & Indo-European & 4 & 10 / 10 \\
Hungarian (hung1274\_Latin) & 69.67 & 71.54 & Uralic & 4 & 10 / 10 \\
Icelandic (icel1247\_Latin) & 65.36 & 69.58 & Indo-European & 2 & 10 / 10 \\
Igbo (nucl1417\_Latin) & 50.62 & 64.71 & Atlantic-Congo & 1 & 10 / 10 \\
Iloko (ilok1237\_Latin) & 56.05 & 69.03 & Austronesian & 1 & 10 / 10 \\
Irish (iris1253\_Latin) & 64.73 & 77.31 & Indo-European & 2 & 10 / 10 \\
Italian (ital1282\_Latin) & 62.85 & 64.59 & Indo-European & 4 & 10 / 10 \\
Japanese (nucl1643\_Japanese) & 53.92 & 72.73 & Japonic & 5 & 10 / 10 \\
Javanese (java1254\_Latin) & 64.70 & 71.11 & Austronesian & 1 & 10 / 10 \\
Kabiyé (kabi1261\_Latin) & 0.44 & 39.03 & Atlantic-Congo & 0 & 3 / 10 \\
Kabuverdianu (kabu1256\_Latin) & 58.01 & 75.68 & Indo-European & -- & 10 / 10 \\
Kabyle (kaby1243\_Latin) & 32.01 & 58.15 & Afro-Asiatic & 1 & 9 / 10 \\
Kamba (Kenya) (kamb1297\_Latin) & 24.93 & 47.68 & Atlantic-Congo & 0 & 8 / 10 \\
Kannada (nucl1305\_Kannada) & 55.88 & 63.90 & Dravidian & 1 & 10 / 10 \\
Kashmiri (kash1277\_Arabic) & 26.62 & 62.82 & Indo-European & 1 & 10 / 10 \\
Kashmiri (kash1277\_Devanagari (Nagari)) & 22.55 & 57.73 & Indo-European & 1 & 10 / 10 \\
Kazakh (kaza1248\_Cyrillic) & 64.98 & 72.01 & Turkic & 3 & 10 / 10 \\
Kikuyu (kiku1240\_Latin) & 5.62 & 53.15 & Atlantic-Congo & 1 & 10 / 10 \\
Kimbundu (kimb1241\_Latin) & 21.37 & 41.79 & Atlantic-Congo & 0 & 8 / 10 \\
Kinshasa Lingala (ling1263\_Latin) & 48.36 & 53.12 & Atlantic-Congo & 1 & 10 / 10 \\
Kinyarwanda (kiny1244\_Latin) & 59.06 & 65.75 & Atlantic-Congo & 1 & 10 / 10 \\
Kirghiz (kirg1245\_Cyrillic) & 54.92 & 59.06 & Turkic & 1 & 10 / 10 \\
Korean (kore1280\_Hangul (Hangŭl, Hangeul)) & 38.21 & 62.31 & Koreanic & 4 & 10 / 10 \\
Lao (laoo1244\_Lao) & 58.59 & 70.02 & Tai-Kadai & 2 & 10 / 10 \\
Levantine Arabic (nort3139\_Arabic) & 67.19 & 74.01 & Afro-Asiatic & -- & 10 / 10 \\
Ligurian (ligu1248\_Latin) & 48.14 & 76.98 & Indo-European & 1 & 10 / 10 \\
Limburgan (limb1263\_Latin) & 56.83 & 76.72 & Indo-European & -- & 10 / 10 \\
Lithuanian (lith1251\_Latin) & 65.99 & 71.04 & Indo-European & 3 & 10 / 10 \\
Lombard (lomb1257\_Latin) & 40.32 & 67.99 & Indo-European & 1 & 10 / 10 \\
Luba-Lulua (luba1249\_Latin) & 29.97 & 52.74 & Atlantic-Congo & 0 & 9 / 10 \\
Luo (Kenya and Tanzania) (luok1236\_Latin) & 37.90 & 47.98 & Nilotic & -- & 10 / 10 \\
Macedonian (mace1250\_Cyrillic) & 64.95 & 70.12 & Indo-European & 1 & 10 / 10 \\
Magahi (maga1260\_Devanagari (Nagari)) & 57.93 & 73.59 & Indo-European & 0 & 10 / 10 \\
Maithili (mait1250\_Devanagari (Nagari)) & 50.43 & 66.99 & Indo-European & 1 & 10 / 10 \\
Malayalam (mala1464\_Malayalam) & 59.07 & 69.10 & Dravidian & 1 & 10 / 10 \\
Maltese (malt1254\_Latin) & 76.21 & 82.70 & Afro-Asiatic & 2 & 10 / 10 \\
Mandarin (mand1415\_Han (Simplified)) & 40.77 & 66.44 & Sino-Tibetan & 5 & 10 / 10 \\
Mandarin (mand1415\_Han (Traditional)) & 34.25 & 68.81 & Sino-Tibetan & 5 & 10 / 10 \\
Manipuri (mani1292\_Bengali) & 19.06 & 64.31 & Sino-Tibetan & 0 & 10 / 10 \\
Maori (maor1246\_Latin) & 47.45 & 64.97 & Austronesian & 1 & 10 / 10 \\
Marathi (mara1378\_Devanagari (Nagari)) & 52.66 & 66.06 & Indo-European & 2 & 10 / 10 \\
Minangkabau (mina1268\_Arabic) & 8.12 & 61.44 & Austronesian & 1 & 10 / 10 \\
Minangkabau (mina1268\_Latin) & 62.69 & 71.41 & Austronesian & 1 & 10 / 10 \\
Mizo (lush1249\_Latin) & 50.39 & 59.40 & Sino-Tibetan & 0 & 10 / 10 \\
Modern Greek (mode1248\_Greek) & 59.10 & 73.07 & Indo-European & 3 & 10 / 10 \\
Modern Hebrew (hebr1245\_Hebrew) & 69.28 & 74.57 & Afro-Asiatic & 3 & 10 / 10 \\
Moroccan Arabic (moro1292\_Arabic) & 45.14 & 60.62 & Afro-Asiatic & 5 & 10 / 10 \\
Moselle Franconian (luxe1241\_Latin) & 59.83 & 75.58 & Indo-European & 1 & 10 / 10 \\
Mossi (moss1236\_Latin) & 15.53 & 40.71 & Atlantic-Congo & 0 & 6 / 10 \\
Najdi Arabic (najd1235\_Arabic) & 65.27 & 72.14 & Afro-Asiatic & -- & 10 / 10 \\
Nepali (nepa1254\_Devanagari (Nagari)) & 52.28 & 70.34 & Indo-European & 1 & 10 / 10 \\
North Azerbaijani (nort2697\_Latin) & 46.62 & 61.61 & Turkic & -- & 10 / 10 \\
Northern Kurdish (nort2641\_Latin) & 46.16 & 64.78 & Indo-European & 0 & 10 / 10 \\
Northern Tosk Albanian (tosk1239\_Latin) & 64.32 & 74.24 & Indo-European & -- & 10 / 10 \\
Northern Uzbek (nort2690\_Latin) & 64.70 & 70.08 & Turkic & -- & 10 / 10 \\
Norwegian Bokmål (norw1259\_Latin) & 67.89 & 70.38 & Indo-European & -- & 10 / 10 \\
Norwegian Nynorsk (norw1262\_Latin) & 68.94 & 77.57 & Indo-European & -- & 10 / 10 \\
Nuer (nuer1246\_Latin) & 6.65 & 21.75 & Nilotic & 0 & 4 / 8 \\
Nyanja (nyan1308\_Latin) & 57.28 & 64.36 & Atlantic-Congo & 1 & 10 / 10 \\
Occitan (occi1239\_Latin) & 64.46 & 75.99 & Indo-European & 1 & 10 / 10 \\
Odia (oriy1255\_Oriya) & 57.08 & 70.09 & Indo-European & 1 & 10 / 10 \\
Pangasinan (pang1290\_Latin) & 50.29 & 67.22 & Austronesian & 1 & 10 / 10 \\
Papiamento (papi1253\_Latin) & 59.40 & 77.99 & Indo-European & 1 & 10 / 10 \\
Pedi (pedi1238\_Latin) & 58.90 & 72.17 & Atlantic-Congo & -- & 10 / 10 \\
Plateau Malagasy (plat1254\_Latin) & 54.33 & 66.30 & Austronesian & 1 & 10 / 10 \\
Polish (poli1260\_Latin) & 63.24 & 68.08 & Indo-European & 4 & 10 / 10 \\
Portuguese (port1283\_Latin) & 74.12 & 72.46 & Indo-European & 4 & 10 / 10 \\
Romanian (roma1327\_Latin) & 72.24 & 73.24 & Indo-European & 3 & 10 / 10 \\
Rundi (rund1242\_Latin) & 46.17 & 59.44 & Atlantic-Congo & 1 & 10 / 10 \\
Russian (russ1263\_Cyrillic) & 70.87 & 69.89 & Indo-European & 4 & 10 / 10 \\
Samoan (samo1305\_Latin) & 52.34 & 70.75 & Austronesian & 1 & 10 / 10 \\
Sango (sang1328\_Latin) & 18.31 & 41.16 & Atlantic-Congo & 1 & 8 / 10 \\
Sanskrit (sans1269\_Devanagari (Nagari)) & 38.77 & 53.26 & Indo-European & 2 & 10 / 10 \\
Santali (sant1410\_Ol Chiki (Ol Cemet’, Ol, Santali)) & 28.85 & 57.77 & Austroasiatic & 1 & 10 / 10 \\
Sardinian (sard1257\_Latin) & 63.26 & 76.16 & Indo-European & 1 & 10 / 10 \\
Scottish Gaelic (scot1245\_Latin) & 56.12 & 68.45 & Indo-European & 1 & 10 / 10 \\
Serbian Standard (serb1264\_Cyrillic) & 63.79 & 74.85 & Indo-European & 4 & 10 / 10 \\
Shan (shan1277\_Myanmar (Burmese)) & 18.45 & 65.01 & Tai-Kadai & 0 & 6 / 10 \\
Shona (shon1251\_Latin) & 50.02 & 53.16 & Atlantic-Congo & 1 & 10 / 10 \\
Sicilian (sici1248\_Latin) & 50.63 & 68.78 & Indo-European & 1 & 10 / 10 \\
Silesian (sile1253\_Latin) & 52.44 & 75.23 & Indo-European & 1 & 10 / 10 \\
Sindhi (sind1272\_Arabic) & 56.57 & 71.45 & Indo-European & 1 & 10 / 10 \\
Sinhala (sinh1246\_Sinhala) & 54.76 & 65.09 & Indo-European & 1 & 10 / 10 \\
Slovak (slov1269\_Latin) & 59.60 & 68.26 & Indo-European & 3 & 10 / 10 \\
Slovenian (slov1268\_Latin) & 70.90 & 72.76 & Indo-European & 3 & 10 / 10 \\
Somali (soma1255\_Latin) & 48.80 & 62.48 & Afro-Asiatic & 1 & 10 / 10 \\
South Azerbaijani (sout2697\_Arabic) & 37.49 & 63.69 & Turkic & -- & 10 / 10 \\
South Levantine Arabic (sout3123\_Arabic) & 53.99 & 70.58 & Afro-Asiatic & -- & 10 / 10 \\
South-Central Koongo (koon1244\_Latin) & 24.58 & 49.15 & Atlantic-Congo & 1 & 8 / 10 \\
Southern Jinghpaw (kach1280\_Latin) & 21.18 & 45.03 & Sino-Tibetan & 0 & 6 / 10 \\
Southern Pashto (sout2649\_Arabic) & 33.63 & 64.12 & Indo-European & -- & 10 / 10 \\
Southern Sotho (sout2807\_Latin) & 55.44 & 75.96 & Atlantic-Congo & 1 & 10 / 10 \\
Southwestern Dinka (sout2832\_Latin) & 1.38 & 24.26 & Nilotic & -- & 4 / 9 \\
Spanish (stan1288\_Latin) & 63.33 & 66.93 & Indo-European & 5 & 10 / 10 \\
Standard Arabic (stan1318\_Arabic) & 67.19 & 71.83 & Afro-Asiatic & 5 & 10 / 10 \\
Standard Arabic (stan1318\_Latin) & 19.46 & 68.76 & Afro-Asiatic & 5 & 10 / 10 \\
Standard Indonesian (indo1316\_Latin) & 74.66 & 69.53 & Austronesian & 3 & 10 / 10 \\
Standard Latvian (stan1325\_Latin) & 63.66 & 73.36 & Indo-European & 3 & 10 / 10 \\
Standard Malay (stan1306\_Latin) & 73.67 & 74.33 & Austronesian & 3 & 10 / 10 \\
Sundanese (sund1252\_Latin) & 53.08 & 60.76 & Austronesian & 1 & 10 / 10 \\
Swahili (swah1253\_Latin) & 75.19 & 77.87 & Atlantic-Congo & 2 & 10 / 10 \\
Swati (swat1243\_Latin) & 47.46 & 59.26 & Atlantic-Congo & 1 & 10 / 10 \\
Swedish (swed1254\_Latin) & 75.76 & 73.53 & Indo-European & 4 & 10 / 10 \\
Ta'izzi-Adeni Arabic (taiz1242\_Arabic) & 57.90 & 68.61 & Afro-Asiatic & -- & 10 / 10 \\
Tagalog (taga1270\_Latin) & 65.38 & 79.03 & Austronesian & 3 & 10 / 10 \\
Tajik (taji1245\_Cyrillic) & 57.78 & 65.02 & Indo-European & 1 & 10 / 10 \\
Tamasheq (tama1365\_Latin) & 12.08 & 35.07 & Afro-Asiatic & 0 & 6 / 10 \\
Tamasheq (tama1365\_Tifinagh (Berber)) & 12.61 & 28.51 & Afro-Asiatic & 0 & 7 / 9 \\
Tamil (tami1289\_Tamil) & 66.37 & 67.33 & Dravidian & 3 & 10 / 10 \\
Tatar (tata1255\_Cyrillic) & 63.39 & 65.85 & Turkic & 1 & 10 / 10 \\
Telugu (telu1262\_Telugu) & 58.70 & 74.29 & Dravidian & 1 & 10 / 10 \\
Thai (thai1261\_Thai) & 64.09 & 74.75 & Tai-Kadai & 3 & 10 / 10 \\
Tibetan (tibe1272\_Tibetan) & 46.95 & 58.32 & Sino-Tibetan & 1 & 10 / 10 \\
Tigrinya (tigr1271\_Ethiopic (Ge`ez)) & 26.43 & 61.36 & Afro-Asiatic & 2 & 10 / 10 \\
Tok Pisin (tokp1240\_Latin) & 46.00 & 58.89 & Indo-European & 1 & 10 / 10 \\
Tsonga (tson1249\_Latin) & 53.82 & 66.83 & Atlantic-Congo & 1 & 10 / 10 \\
Tswana (tswa1253\_Latin) & 45.34 & 62.99 & Atlantic-Congo & 2 & 10 / 10 \\
Tumbuka (tumb1250\_Latin) & 48.32 & 58.45 & Atlantic-Congo & 1 & 10 / 10 \\
Tunisian Arabic (tuni1259\_Arabic) & 43.91 & 67.20 & Afro-Asiatic & -- & 10 / 10 \\
Turkish (nucl1301\_Latin) & 69.30 & 78.82 & Turkic & 4 & 10 / 10 \\
Turkmen (turk1304\_Latin) & 54.86 & 67.57 & Turkic & 1 & 10 / 10 \\
Twi (twii1234\_Latin) & 40.08 & 54.68 & Atlantic-Congo & 1 & 10 / 10 \\
Uighur (uigh1240\_Arabic) & 57.10 & 63.85 & Turkic & 1 & 10 / 10 \\
Ukrainian (ukra1253\_Cyrillic) & 67.63 & 73.64 & Indo-European & 3 & 10 / 10 \\
Umbundu (umbu1257\_Latin) & 19.95 & 44.89 & Atlantic-Congo & 0 & 7 / 10 \\
Urdu (urdu1245\_Arabic) & 56.80 & 69.39 & Indo-European & 3 & 10 / 10 \\
Venetian (vene1258\_Latin) & 53.60 & 72.88 & Indo-European & 1 & 10 / 10 \\
Vietnamese (viet1252\_Latin) & 68.50 & 67.29 & Austroasiatic & 4 & 10 / 10 \\
Waray (Philippines) (wara1300\_Latin) & 61.97 & 80.62 & Austronesian & 1 & 10 / 10 \\
Welsh (wels1247\_Latin) & 76.84 & 80.79 & Indo-European & 1 & 10 / 10 \\
West Central Oromo (west2721\_Latin) & 43.92 & 58.33 & Afro-Asiatic & -- & 10 / 10 \\
Western Farsi (west2369\_Arabic) & 51.22 & 69.55 & Indo-European & -- & 10 / 10 \\
Wolof (nucl1347\_Latin) & 27.23 & 52.05 & Atlantic-Congo & 2 & 9 / 10 \\
Xhosa (xhos1239\_Latin) & 51.60 & 64.15 & Atlantic-Congo & 2 & 10 / 10 \\
Yoruba (yoru1245\_Latin) & 25.90 & 50.06 & Atlantic-Congo & 2 & 10 / 10 \\
Yue Chinese (yuec1235\_Han (Traditional)) & 30.09 & 68.45 & Sino-Tibetan & 1 & 10 / 10 \\
Zulu (zulu1248\_Latin) & 58.61 & 74.58 & Atlantic-Congo & 2 & 10 / 10 \\
\end{longtable}
}

\clearpage

\section{Scores to be Compared with the Grammar Agent Setting}

\begin{table}[htbp]
\centering
\resizebox{\textwidth}{!}{%
\begin{tabular}{lrrrrrr}
\toprule
\textbf{Run ID} & \textbf{Avg Score} & \textbf{Scores $\boldsymbol{<0.25}$} & \textbf{Scores $\boldsymbol{[0.25, 0.5)}$} & \textbf{Scores $\boldsymbol{[0.5, 0.75)}$} & \textbf{Scores $\boldsymbol{\ge 0.75}$} & \textbf{Total} \\
\midrule

OpenAI-o4-mini (baseline) & 0.193 & 50 & 15 & 3 & 5 & 73 \\
Gemini-2.5-pro (baseline) & 0.373 & 33 & 19 & 13 & 12 & 77 \\
OpenAI-o4-mini (MoA, 2 rounds) & 0.319 & 35 & 20 & 10 & 10 & 75 \\
Gemini-2.5-pro (MoA, 2 rounds) & 0.409 & 26 & 20 & 18 & 12 & 76 \\
OpenAI-o4-mini (MoA, 3 rounds) & 0.337 & 35 & 18 & 15 & 9 & 77 \\
Gemini-2.5-pro (MoA, 3 rounds) & 0.416 & 27 & 18 & 20 & 12 & 77 \\
\midrule
\bottomrule
\end{tabular}%
}
\caption{The baseline and Mixture-of-Agents scores to be compared with the system with Grammar Agent. Because only a portion of problems have reference grammar books, and it's likely that only the more common language has resources available, we filtered the problems to be the same the when comparing with it.}
\label{tab:agent_performance_summary_subset}
\end{table}

\clearpage

\section{Supplementary Figures}
\label{appendix:supp-figures}

\begin{figure}[htbp]
    \centering
    \includegraphics[width=\textwidth]{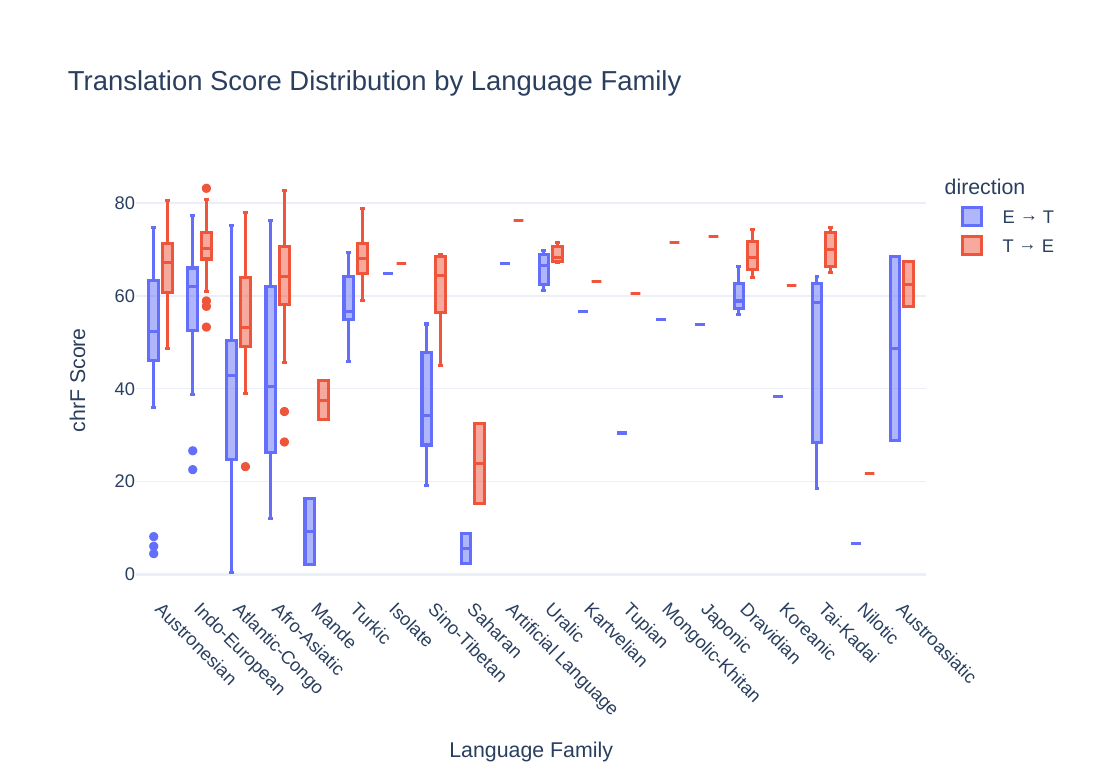}
    \caption{
        \textbf{Translation Score Distribution by Language Family.}
        This plot compares the distribution of \textsc{chrF} scores for English-to-Target ($E \rightarrow T$) and Target-to-English ($T \rightarrow E$) directions across language families. A consistent performance gap is evident, with $T \rightarrow E$ scores being almost universally higher and often less variable than $E \rightarrow T$ scores. Families such as Saharan and Mande show particularly low performance in the $E \rightarrow T$ direction, whereas families like Indo-European show a wider range of performance with generally higher scores.
    }
    \label{fig:chrf_by_family}
\end{figure}

\clearpage

\begin{figure}[htbp]
    \centering
    \includegraphics[width=\textwidth]{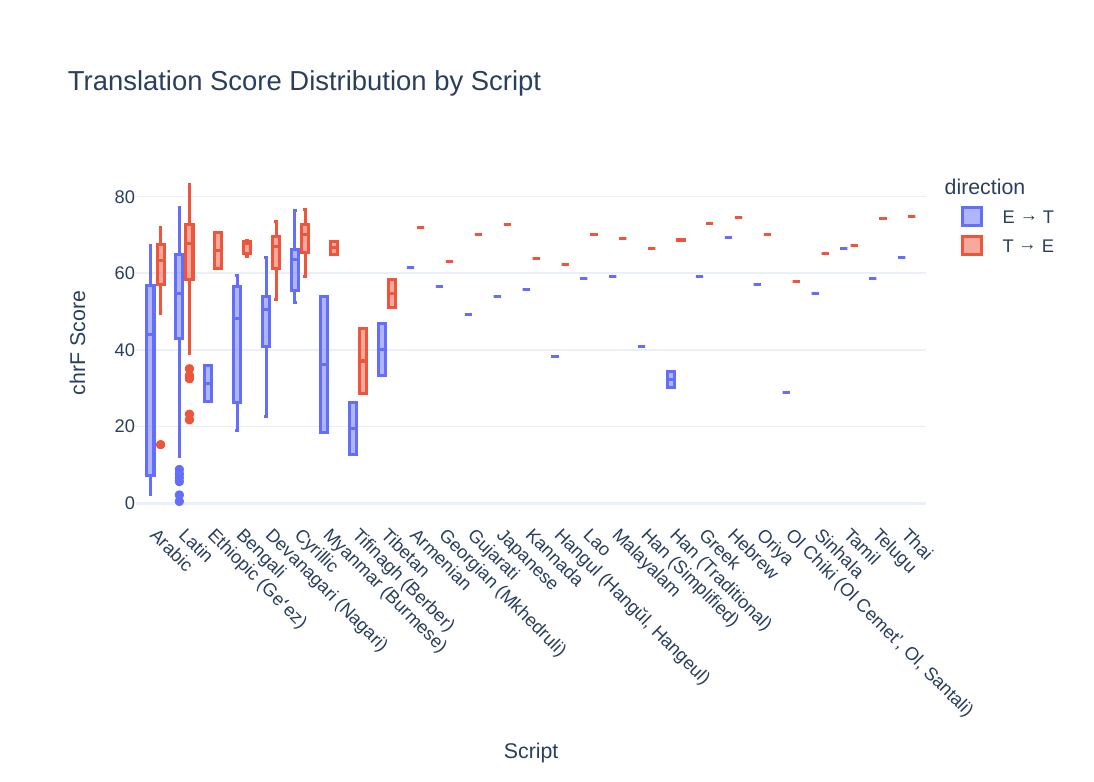}
    \caption{
        \textbf{Translation Score Distribution by Script.}
        This plot compares \textsc{chrF} score distributions across different writing systems. As with the family-based plot, the $T \rightarrow E$ direction consistently outperforms the $E \rightarrow T$ direction. Performance for languages using Latin and Cyrillic scripts is relatively high but shows a wide distribution, reflecting the diverse range of languages using them. Scripts associated with lower-resource languages, such as Ethiopic and Tifinagh, exhibit lower median scores, particularly in the $E \rightarrow T$ direction.
    }
    \label{fig:chrf_by_script}
\end{figure}

\clearpage

\begin{figure}[htbp]
    \centering
    \includegraphics[width=\textwidth]{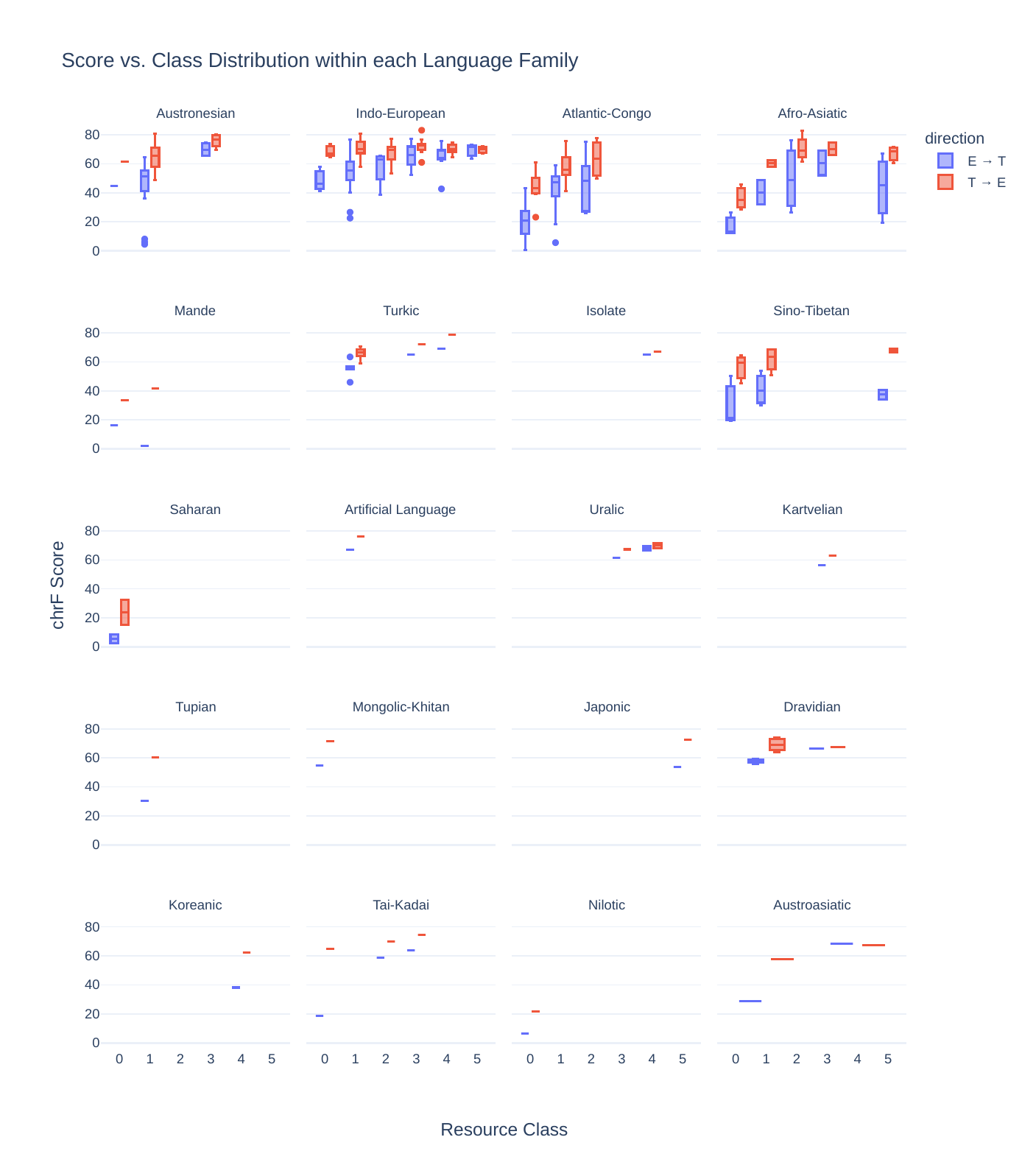}
    \caption{
        \textbf{Score vs. Class Distribution within each Language Family.}
        This faceted plot details the relationship between resource class and \textsc{chrF} score for each language family individually. A positive trend, where higher scores are associated with higher resource classes, is visible within several major families like Indo-European and Afro-Asiatic. The plot also highlights data sparsity, as many families (e.g., Mande, Saharan, Nilotic) contain languages in only one or two resource classes. The performance gap between the two translation directions persists even when controlling for class within a family.
    }
    \label{fig:faceted_by_family}
\end{figure}

\clearpage

\begin{figure}[htbp]
    \centering
    \includegraphics[width=\textwidth]{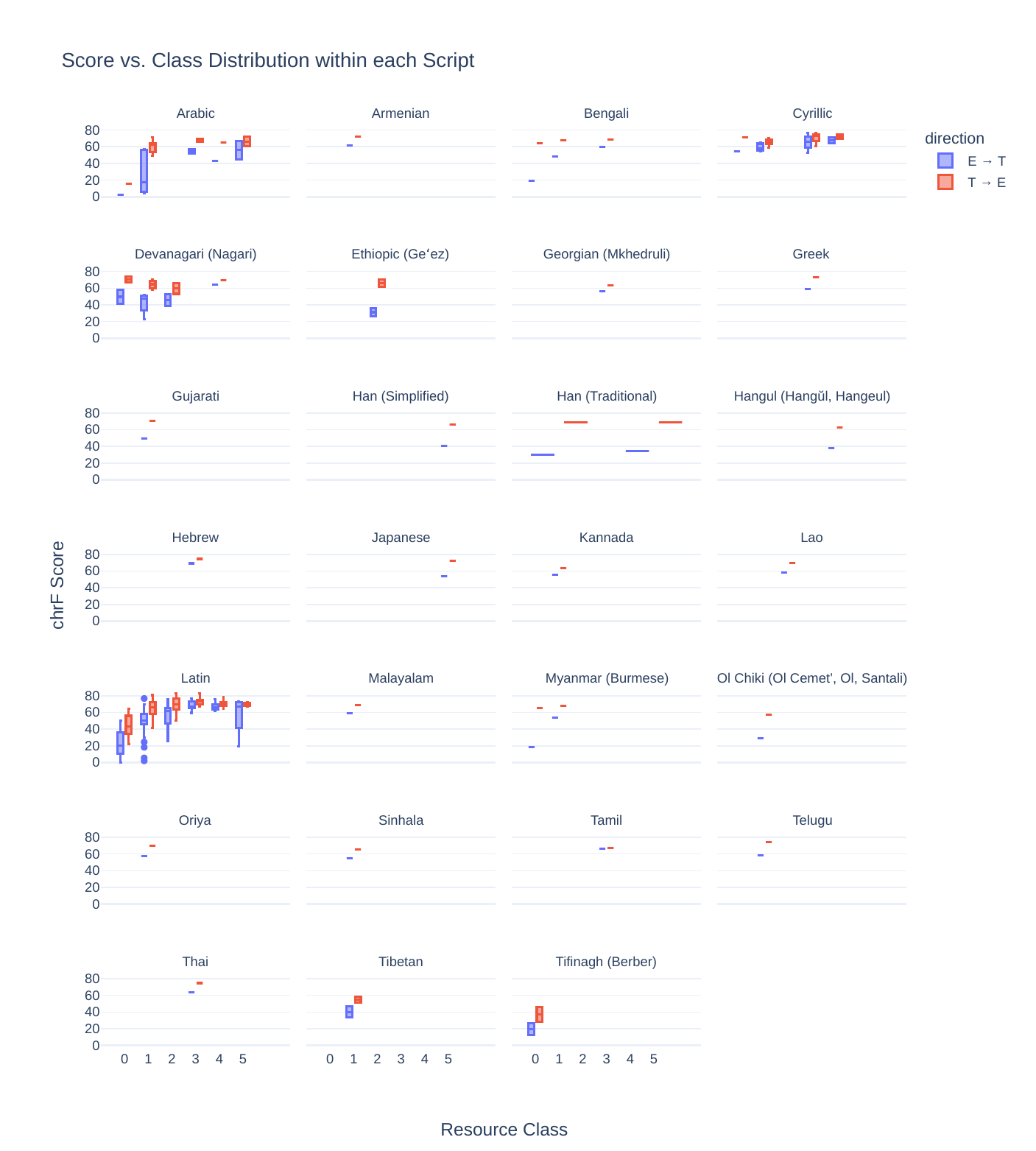}
    \caption{
        \textbf{Score vs. Class Distribution within each Script.}
        This faceted plot shows the relationship between resource class and \textsc{chrF} score for each writing system. The Latin script subplot contains the most data across all resource classes and most clearly demonstrates the positive correlation between class and score. For many other scripts, such as Arabic and Devanagari, the data is concentrated in the lower resource classes. This visualization confirms that the relationship between script and score is highly confounded with resource availability.
    }
    \label{fig:faceted_by_script}
\end{figure}

\end{document}